\documentclass{article}

\usepackage[T1]{fontenc}
\usepackage[utf8]{inputenc}
\usepackage{microtype}

\usepackage{graphicx}
\usepackage{mathtools}
\usepackage{booktabs}
\usepackage{array}
\usepackage{tabularx}
\usepackage{longtable}
\usepackage{makecell}
\usepackage{multirow}
\usepackage{threeparttable}
\usepackage{amsmath}
\usepackage{amssymb}

\usepackage[table,xcdraw]{xcolor}

\usepackage{tikz}
\usetikzlibrary{shapes,arrows.meta,positioning,calc}

\usepackage{svg}
\svgpath{{figures/}}

\usepackage{float}
\usepackage{placeins}
\usepackage{dblfloatfix}
\usepackage{pdflscape}

\usepackage{enumitem}
\usepackage{etoolbox}

\usepackage[
    colorlinks=true,
    urlcolor=blue,
    linkcolor=black,
    citecolor=black
]{hyperref}

\usepackage[accepted]{icml2025}

\makeatletter
\renewcommand{\Notice@String}{Preprint.}
\makeatother

\icmltitlerunning{
A Comparative Assessment of Global Building and Settlement Datasets
}

\begin{document}

\twocolumn[

\icmltitle{
A Comparative Assessment of Global Building and Settlement Datasets
Across Geographic and Settlement Contexts
}


\begin{icmlauthorlist}
\icmlauthor{Rufai Omowunmi Balogun}{wb,clark}
\icmlauthor{Caroline Margaux Gevaert}{wb,utwente}
\icmlauthor{Capucine Riom}{wb}
\icmlauthor{Derrick Mirindi}{wb,penn,morgan}
\icmlauthor{Aaron Opdyke}{wb}
\icmlauthor{Hamed Alemohammad}{clark}
\icmlauthor{Pierre Chrzanowski}{wb}
\icmlauthor{Edward Charles Anderson}{wb}
\end{icmlauthorlist}

\icmlcorrespondingauthor{Rufai Omowunmi Balogun}{rbalogun@clarku.edu}
\icmlcorrespondingauthor{Caroline Margaux Gevaert}{cgevaert@worldbank.org }

\icmlaffiliation{wb}{
The World Bank,
Global Facility for Disaster Reduction and Recovery
}

\icmlaffiliation{utwente}{
University of Twente,
Enschede, The Netherlands
}

\icmlaffiliation{clark}{
Center for Geospatial Analytics,
Clark University,
Worcester, Massachusetts, USA
}

\icmlaffiliation{penn}{
School of Engineering and Applied Science,
University of Pennsylvania,
Philadelphia, Pennsylvania, USA
}

\icmlaffiliation{morgan}{
School of Architecture and Planning,
Morgan State University,
Baltimore, Maryland, USA
}


\vskip 0.3in
]

\printAffiliationsAndNotice{}


\begin{abstract}

Global building and settlement products increasingly support population mapping, exposure assessment, and urban monitoring, yet comparative evidence remains fragmented across representations, geographic contexts, and evaluation methods. We benchmark seven global or near-global products against harmonized reference building footprints across 135 study areas. Our representation-aware framework integrates complementary measures of detection, geometric agreement, and aggregate quantity accuracy with stratified analyses of settlement characteristics, evaluation resolution, reference provenance, and temporal alignment. Overture Maps achieved the highest median city-level vector F1 (0.786), while 3D-GloBFP showed the closest aggregate building-area agreement despite weaker detection performance. Raster rankings were scale-dependent: Google Open Buildings Temporal (OBT) performed best at 10\,m (F1 = 0.680), whereas WSF Tracker led at 100\,m (F1 = 0.904) but substantially overestimated reference building area (relative bias = 1.845). Raster accuracy increased with building density ($\rho=0.58$--$0.75$), while smaller candidate footprints were disproportionately associated with vector false positives. Temporal alignment improved WSF Tracker mean city-level F1 by 0.060. These findings demonstrate that product reliability depends on representation, spatial scale, and settlement context, providing a reproducible framework for assessing the suitability of EO-derived urban information for different applications.



\end{abstract}


\section{Introduction}
\label{sec:introduction}

\begin{figure*}
\centering
\includegraphics[width=\linewidth]{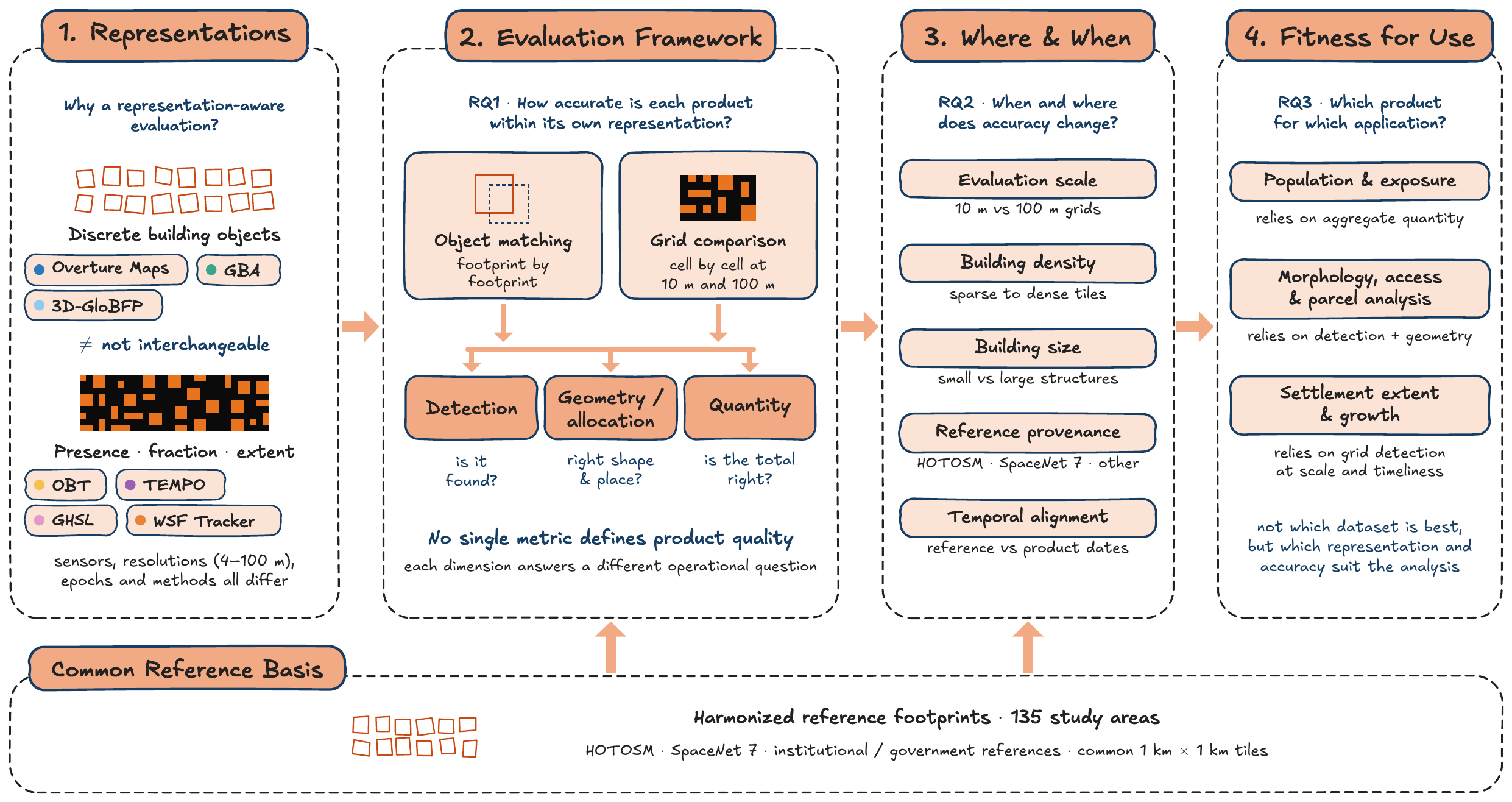}
\caption{Overview of the representation-aware evaluation framework. Vector and raster building products are evaluated within their respective representations against a common reference dataset. Accuracy is assessed in terms of detection, geometric agreement, and quantity, and examined across settlement, spatial-scale, reference, and temporal contexts to inform fitness for use. Detailed workflows are described in Section~\ref{sec:methodology}.}
\label{fig:conceptual_rep_aware}
\end{figure*}

Earth observation (EO) has become an invaluable source of independent information for characterizing the spatial properties of the built environment and inferring other critical asset-level information. These data provide important context and serve as critical inputs for applications such as population mapping \citep{lloyd2019,boo2022}, exposure and disaster risk modelling \citep{yepes2023,GUNASEKERA2015594}, urban growth monitoring \citep{pesaresi2024ghsl}, infrastructure access, and urban informality \citep{bettencourt2025infrastructure}. Advances in satellite-based optical and radar imaging, machine learning, volunteered geographic information, and large-scale EO analytics have enabled the production of building and settlement datasets at increasingly broad geographic scales, from national to regional and global coverage. These developments are particularly valuable in areas with limited authoritative cadastral or municipal data, especially in the Global South, where such information supports critical planning and policy decisions \citep{unhabitat2022world}.

Despite their widespread utility, the proliferation of global building and settlement products presents an important challenge for urban EO. These products differ in what they measure and how they represent information, drawing on different sensor modalities, observation periods, spatial resolutions, and analytical methods. Vector products typically delineate individual building footprints, whereas raster products may represent building presence, built-up fraction, building density, height, or broader settlement extent \citep{zhu2025gba,pesaresi2024ghsl,glazer2025tempo}. Consequently, these products encode different abstractions of the built environment rather than interchangeable estimates of a single quantity. Although this diversity reflects varying operational needs, it complicates rigorous cross-product evaluation. Meaningful comparisons must therefore account for differences in representation and spatial scale while providing a consistent basis for assessing performance.

Existing validation approaches generally follow two practices: (i) evaluating individual-building detection and geometric agreement \citep{chamberlain2024africa,okyere2025obd,sirko2021continental,zhu2025gba}; and (ii) assessing grid-level agreement and building attributes or quantities, such as height, built-up area, and estimated building counts \citep{marconcini2020wsf,pesaresi2024ghsl,liu2020ghsl}. These approaches typically follow the inherent vector or raster representation of the evaluated products and report aggregate accuracy scores. However, such scores can conceal systematic variation in performance, particularly where buildings are small, densely clustered, or difficult to resolve. Previous studies have therefore examined accuracy variation across geographic and socioeconomic contexts \citep{gevaert2024,liu2020ghsl,lahrichi2026buildingarea}. Despite these advances, evaluations remain fragmented, with few studies jointly assessing vector and raster products while accounting for their distinct representations. Furthermore, limited attention has been given to how performance varies with settlement morphology, spatial scale, reference provenance, and temporal alignment. These gaps constrain our ability to assess product reliability, geographic robustness, and suitability for different urban applications.

In this study, we address these gaps by introducing a representation-aware framework for jointly evaluating vector and raster building and settlement products across diverse geographic contexts. We evaluate seven global or near-global products against harmonized reference building footprints collected from 135 cities, communities, and municipalities, with substantial representation from the Global South. The evaluated vector products are Overture Maps Buildings, the Global Building Atlas (GBA), and 3D-GloBFP \citep{overture2026buildings,zhu2025gba,che20243dglobfp}. The raster products comprise Google Open Buildings 2.5D Temporal (OBT), Microsoft TEMPO, the Global Human Settlement Layer (GHSL), and the World Settlement Footprint (WSF) Tracker \citep{sirko2023sentinel2,glazer2025tempo,pesaresi2024ghsl,dlr2026wsftracker}. Specifically, we address three research questions: (1) How accurately do current global vector and raster products represent the built environment within their respective representations? (2) How does performance vary with evaluation resolution, building density and size, reference-data provenance, and temporal alignment? (3) How do detection, geometric agreement, and aggregate quantity agreement affect the interpretation and suitability of these products for different urban applications?

The study contributes to EO-based urban analytics in three ways. First, it establishes a harmonized multi-source benchmark for seven global building and settlement products across 135 diverse study areas (Section~\ref{sec:datasets}). Second, it introduces a representation-aware evaluation framework linking object-level, grid-level, geometric, and aggregate measures to the properties of the built environment represented by these products (Section~\ref{sec:methodology}). Figure~\ref{fig:conceptual_rep_aware} summarizes the framework and its principal evaluation dimensions. Third, the study combines overall benchmarking with stratified and diagnostic analyses to identify how settlement morphology, evaluation scale, reference provenance, and temporal alignment influence performance (Section~\ref{subsec:context_variation}). Together, these contributions provide an open and reproducible framework for assessing the reliability and suitability of global urban products across data representations, spatial scales, settlement contexts, and analytical needs.

\section{Global Building and Settlement Datasets}
\label{sec:datasets}

\begin{figure*}[!t]
  \centering
  \includegraphics[width=\linewidth]{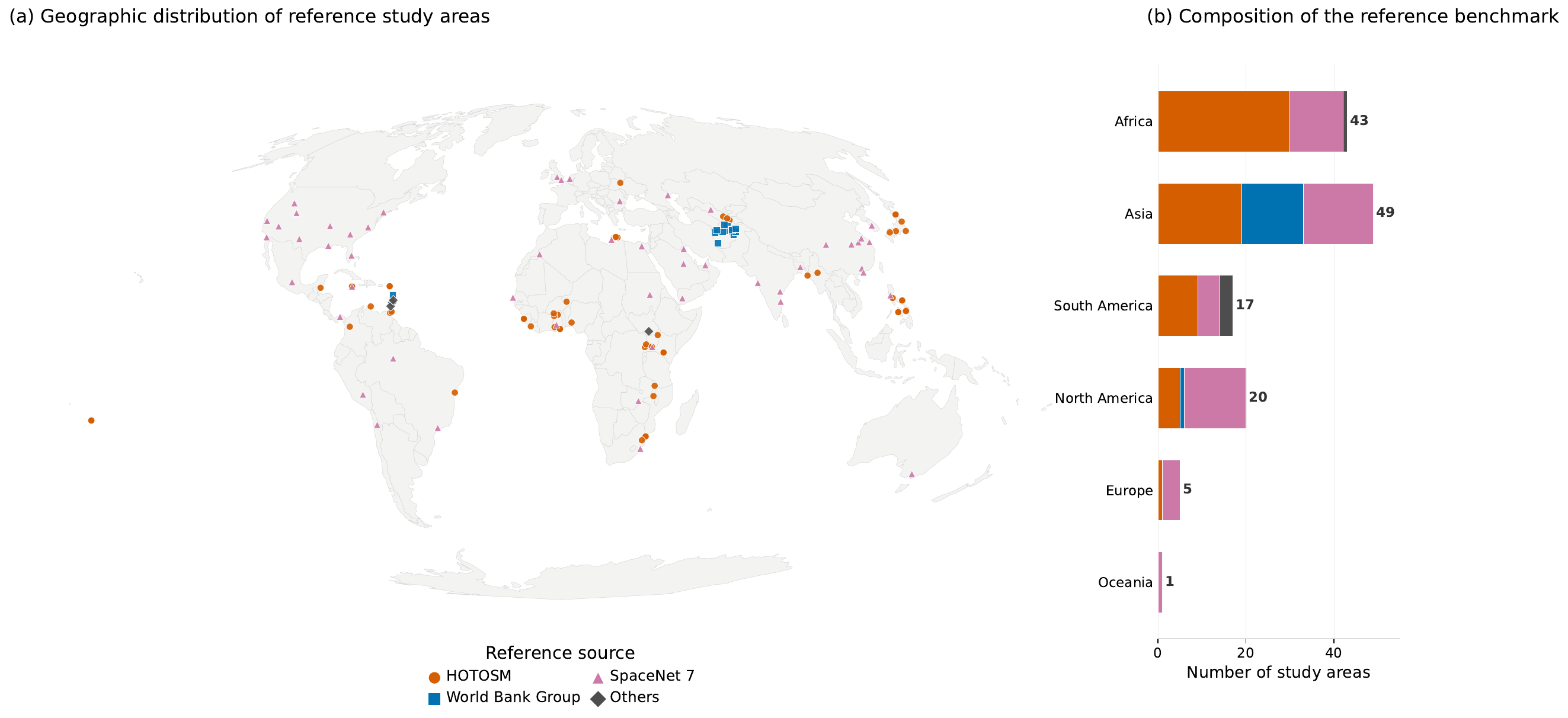}
  \caption{Geographic distribution and composition of the reference benchmark. (a) Locations of the 135 study areas, classified by reference-data source. (b) Number of study areas by geographic region and reference source.}
  \label{fig:ref_datasets}
\end{figure*}

This study evaluates seven global or near-global building and settlement products against harmonized reference building footprints from 135 study areas, with substantial representation of the Global South (Figure~\ref{fig:ref_datasets}). The evaluated products comprise three vector building-footprint datasets and four raster products representing different properties of the built environment. Their spatial and temporal characteristics are summarized in Table~\ref{tab:dataset_comparison}.

\begin{table*}[!t]
\centering
\begin{threeparttable}

\caption{Description of the spatiotemporal characteristics of the global building footprints and settlement layers evaluated in this study, sourced from publicly available datasets. Temporal coverage denotes the period represented by the published product, source imagery epoch denotes the acquisition period of the underlying input imagery, and latest update denotes the most recent verified public release.}
\label{tab:dataset_comparison}

\renewcommand{\arraystretch}{1.25}
\setlength{\tabcolsep}{3pt}
\scriptsize

\begin{tabularx}{\textwidth}{
@{}
>{\raggedright\arraybackslash}p{3.0cm}
>{\raggedright\arraybackslash}p{2.25cm}
>{\raggedright\arraybackslash}p{2.25cm}
>{\raggedright\arraybackslash}p{2.25cm}
>{\raggedright\arraybackslash}p{2.25cm}
>{\raggedright\arraybackslash}p{2.25cm}
>{\raggedright\arraybackslash}p{2.25cm}
@{}
}
\toprule
\textbf{Dataset}
& \textbf{Spatial coverage}
& \textbf{Temporal coverage}
& \textbf{Update cadence}
& \textbf{Spatial resolution}
& \textbf{Source imagery epoch}
& \textbf{Latest update} \\
\midrule

\multicolumn{7}{@{}l}{\textit{Polygon / vector building footprints}} \\
\midrule

Overture Maps
& Global
& Current composite
& Monthly
& Polygons
& Mixed epochs
& Jul 2026 \\

Global Building Atlas (GBA)
& Global
& Static; mixed epochs
& Static
& Polygons
& 2019 primary epoch
& Sep 2025 \\

3D-GloBFP
& Global
& 2020
& Static
& Polygons
& Primarily 2020
& May 2025 \\

\midrule
\multicolumn{7}{@{}l}{\textit{Raster settlement / building products}} \\
\midrule

Google Open Buildings 2.5D Temporal
& Global South / selected regions
& 2016--2023
& Annual
& 4\,m effective
& 2016--2023
& Sep 2024 \\

Microsoft TEMPO
& Global
& 2018--2025
& Quarterly
& 37.6\,m;100\,m
& Q1 2018--Q2 2025
& May 2026 \\

GHSL 
& Global
& 1975--2030
& 5-year epochs
& 100\,m; 10\,m (2018)
& 1975--2018**
& May 2023 \\

WSF Tracker
& Global
& Jul 2016--Jan 2026
& 6-monthly
& 10\,m
& Jul 2016--Jan 2026
& Jun 2026 \\

\bottomrule
\end{tabularx}

\begin{tablenotes}[flushleft]
\footnotesize
\item \textit{Source imagery:}
\textbf{GBA} primarily uses PlanetScope imagery for additional building extraction and incorporates fused third-party building sources with heterogeneous epochs.
\textbf{3D-GloBFP} combines building-footprint and Earth-observation inputs centred approximately on 2020.
\textbf{Google Open Buildings 2.5D Temporal} is derived from Sentinel-2 imagery.
\textbf{TEMPO} is derived from PlanetScope imagery; its 37.6\,m products are available only for selected areas, while the global 100\,m product is available for Q4 2020 and Q4 2023.
\textbf{GHSL (GHS-BUILT-S R2023A)} uses Landsat for the historical observed epochs and Sentinel-2 for the 2018 epoch. The observed imagery epochs are 1975, 1990, 2000, 2014, and 2018; additional layers are provided for projected epochs in 2025 and 2030.
\textbf{WSF Tracker} is derived from Sentinel-1 and Sentinel-2 imagery.
\textbf{Overture Maps} combines multiple upstream building sources and therefore has no single underlying imagery source or epoch.
\end{tablenotes}

\end{threeparttable}
\end{table*}

The \textit{vector products} delineate individual building footprints but differ in their source data and production methods. \textbf{Overture Maps} integrates community-contributed and machine-learning-derived footprints from multiple sources, resulting in heterogeneous observation dates \citep{linuxfoundation2022overture,overture2026buildings}. The \textbf{Global Building Atlas (GBA)} contains over 2.75 billion building polygons with associated height estimates, produced through the fusion of existing building sources and PlanetScope-based extraction \citep{zhu2025gba}. \textbf{3D-GloBFP} provides global building footprints and height estimates centered approximately on 2020, derived from multisource EO data and building-morphology features \citep{che20243dglobfp}.

The \textit{raster products} represent different building and settlement properties at varying spatial and temporal resolutions. \textbf{Google Open Buildings 2.5D Temporal (OBT)} provides annual estimates of building presence, fractional building count, and height from Sentinel-2 imagery \citep{sirko2023sentinel2}. \textbf{Microsoft TEMPO} estimates building density and height from PlanetScope imagery at quarterly intervals \citep{glazer2025tempo}. The \textbf{Global Human Settlement Layer (GHSL)} GHS-BUILT-S R2023A provides multitemporal estimates of built-up surface, including historical observations and projected epochs \citep{pesaresi2023ghsbuilts,pesaresi2024ghsl}. The \textbf{World Settlement Footprint (WSF) Tracker} maps settlement extent at approximately six-month intervals using Sentinel-1 and Sentinel-2 imagery \citep{marconcini2020wsf,marconcini2021wsfsuite,esch2022wsf3d}.

The \textit{reference benchmark} combines building footprints from four sources: (1) \textbf{Humanitarian OpenStreetMap Team (HOTOSM)} mapping campaigns based on high-resolution imagery \citep{herfort2021evolution,firth2020hot}; (2) \textbf{World Bank-supported projects} providing building inventories for urban development, infrastructure, and disaster-risk applications; (3) \textbf{SpaceNet~7}, comprising manually delineated building polygons from multitemporal imagery across 52 cities, organized as $1024\times1024$ image chips \citep{vanetten2021dataset}; and (4) \textbf{University of Edinburgh} reference footprints digitized at 1:200 scale from February 2018 aerial imagery of informal settlements in Cape Town. Reference data were screened for geometric quality, temporal suitability, and consistency with visible buildings. Together, these sources provide geographically and morphologically diverse reference data while preserving source provenance and, where available, imagery acquisition dates for subsequent diagnostic analyses.

\section{Representation-Aware Evaluation Framework}
\label{sec:methodology}

The proposed framework jointly evaluates vector and raster building and settlement products while preserving their distinct representations. It follows three principles: (i) representation-specific evaluation, using object-based matching for vector products and grid-based comparison for raster products; (ii) complementary assessment of detection, geometric or spatial agreement, and aggregate quantity accuracy; and (iii) stratified and diagnostic analyses of performance across settlement characteristics, spatial scales, reference provenance, and temporal alignment. The framework enables consistent benchmarking without treating different representations or evaluation scales as directly interchangeable.

\subsection{Spatial Harmonization and Tiling}
\label{subsec:preprocessing_reference}

Candidate and reference datasets were clipped to each area of interest (AOI), reprojected to the local UTM zone, and screened for invalid geometries. Polygons smaller than $\tau_{\mathrm{size}}=20\,\mathrm{m}^2$ were excluded to reduce digitization artefacts, where $\tau_{\mathrm{size}}$ is the minimum retained footprint area. All spatial measurements and matching operations were performed in the projected coordinate system.

Evaluation used fixed $1\,\mathrm{km}\times1\,\mathrm{km}$ tiles as common units for accuracy assessment and contextual stratification. Reference buildings were assigned to tiles by their centroids to prevent double-counting in building-count, density, and mean-size calculations. Tile-level metrics were aggregated to city level, and benchmark-level performance was summarized using unweighted city medians to prevent larger building inventories from dominating comparisons.

\subsection{Vector Building-Footprint Evaluation}
\label{subsec:vector_evaluation}

\begin{figure*}[!t]
\centering
\includegraphics[width=\linewidth]{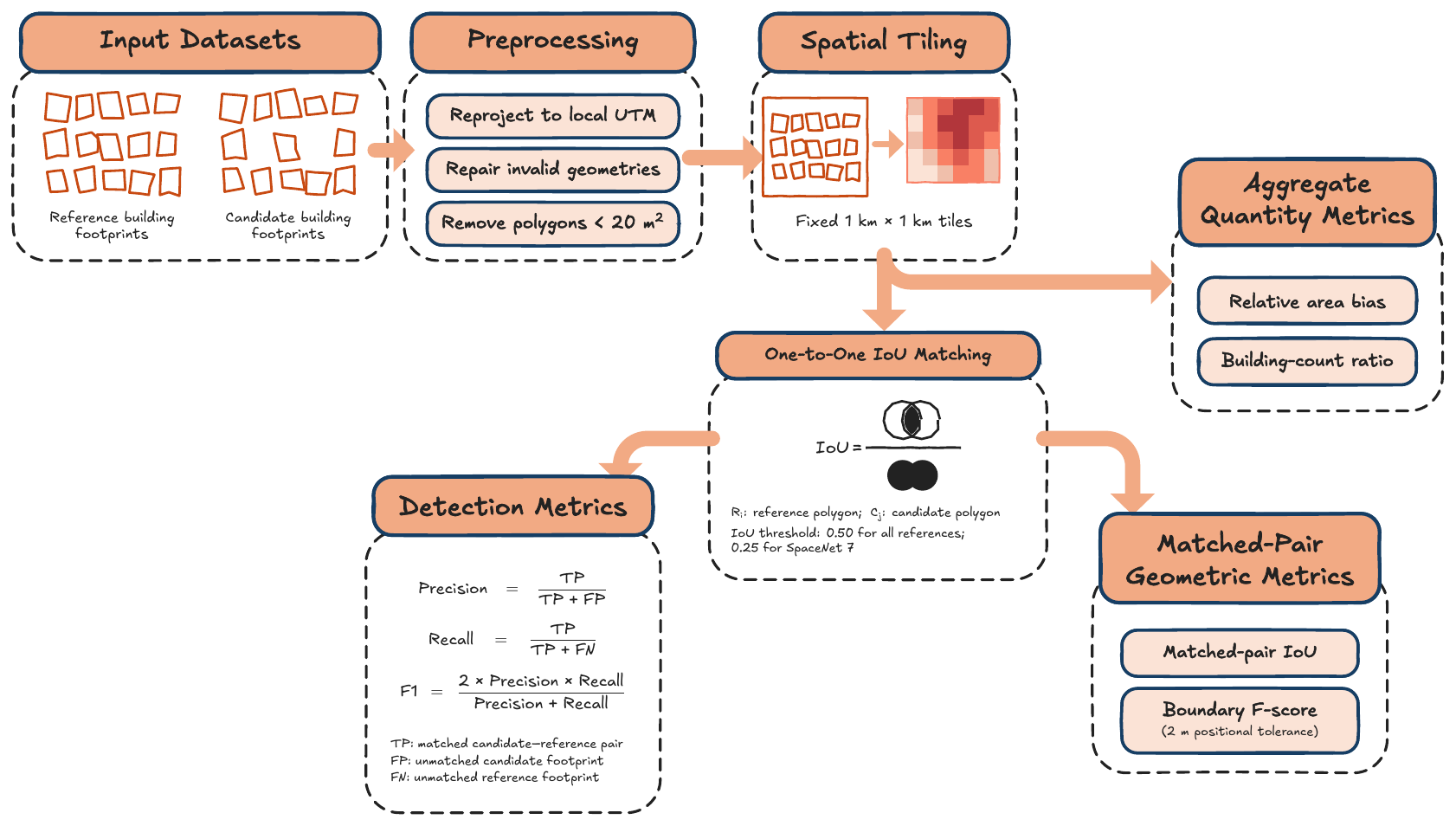}
\caption{Vector evaluation workflow. Candidate and reference footprints are harmonized within 1\,km $\times$ 1\,km tiles. One-to-one IoU matching supports object-detection and geometric assessment, while building-area and count agreement are evaluated independently.}
\label{fig:validation-pipeline}
\end{figure*}

Vector products were assessed through three complementary dimensions: object detection, matched-footprint geometry, and aggregate building quantities (Figure~\ref{fig:validation-pipeline}).

\noindent\textbf{Object detection and geometric agreement.}
Candidate and reference buildings were matched using intersection over union (IoU):

\begin{equation}
\operatorname{IoU}(R_i,C_j)
=
\frac{|R_i\cap C_j|}{|R_i\cup C_j|},
\label{eq:iou}
\end{equation}

where $R_i$ and $C_j$ denote reference building $i$ and candidate building $j$, respectively; $|R_i\cap C_j|$ and $|R_i\cup C_j|$ are their intersection and union areas. Greedy one-to-one matching retained the highest-IoU candidate for each reference building, subject to $\operatorname{IoU}\geq\tau_{\mathrm{overlap}}$, where $\tau_{\mathrm{overlap}}=0.50$, except for SpaceNet~7 ($0.25$), following its evaluation convention \citep{vanetten2021challenge}.

Matched pairs were classified as true positives (TP), unmatched candidates as false positives (FP), and unmatched references as false negatives (FN). Detection accuracy was summarized using precision, recall, and F1 score. For matched pairs, geometric agreement was assessed using IoU and boundary F-score with a positional tolerance of $\tau_{\mathrm{misalign}}=2\,\mathrm{m}$, where $\tau_{\mathrm{misalign}}$ is the maximum permitted boundary displacement.

\medskip
\noindent\textbf{Aggregate quantity agreement.}
To distinguish individual-building detection from agreement in total building stock, relative building-area bias and building-count ratio were calculated independently:

\begin{equation}
B_A =
\frac{A_{\mathrm{pred}}-A_{\mathrm{ref}}}
     {A_{\mathrm{ref}}},
\label{eq:area_bias}
\end{equation}

where $B_A$ is relative area bias, and $A_{\mathrm{pred}}$ and $A_{\mathrm{ref}}$ are the total candidate and reference building areas within the evaluation unit, respectively. Positive and negative values indicate overestimation and underestimation, while zero indicates exact area agreement.

\begin{equation}
R_n =
\frac{n_{\mathrm{pred}}}{n_{\mathrm{ref}}},
\label{eq:count_ratio}
\end{equation}

where $R_n$ is the building-count ratio, and $n_{\mathrm{pred}}$ and $n_{\mathrm{ref}}$ are the candidate and reference building counts, respectively. Values above or below one indicate overcounting or undercounting.

\subsection{Raster Settlement-Layer Evaluation}
\label{subsec:raster_evaluation}

\begin{figure*}[!t]
\centering
\includegraphics[width=\linewidth]{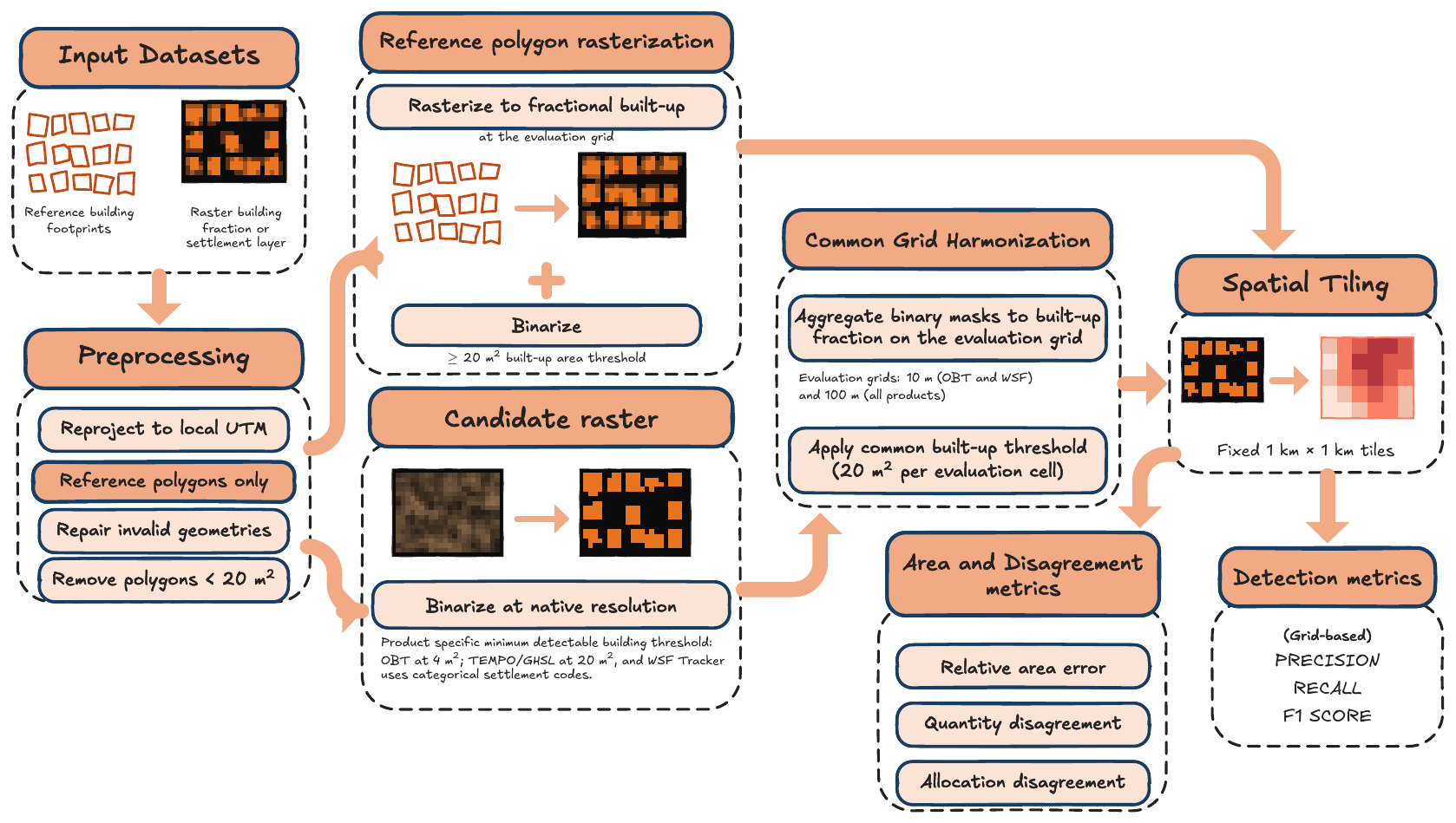}
\caption{Raster evaluation workflow. Reference footprints are rasterized to built-up coverage, while candidate products are classified at native resolution and aggregated to common 10\,m or 100\,m evaluation grids. Grid-level detection, area bias, and quantity/allocation disagreement are assessed within 1\,km tiles.}
\label{fig:raster-validation-pipeline}
\end{figure*}

Raster products were evaluated through scale-aware harmonization, preserving their native spatial support before comparison on common evaluation grids (Figure~\ref{fig:raster-validation-pipeline}). The workflow comprised three stages.

\noindent\textbf{Reference rasterization.}
Reference footprints were rasterized to fractional built-up coverage at each evaluation resolution. Cells were classified as built-up when covered by at least $20\,\mathrm{m}^2$ of reference footprints, equivalent to fractional thresholds of 0.20 at 10\,m and 0.002 at 100\,m.

\medskip
\noindent\textbf{Native-resolution classification.}
Continuous candidate products were binarized using product-specific minimum built-up-area thresholds. OBT used $4\,\mathrm{m}^2$ (approximately 0.25 of a 4\,m pixel), while TEMPO and GHSL used $20\,\mathrm{m}^2$ (0.002 at 100\,m). Categorical WSF Tracker layers were classified directly from settlement codes. Product-specific parameters are provided in Supplementary materials ~\ref{app:raster_binarisation} and Table~\ref{tab:raster_thresholds}.

\medskip
\noindent\textbf{Common-grid harmonization.}
Native binary masks were block-aggregated and reclassified on the evaluation grid using the same $20\,\mathrm{m}^2$ built-up-area criterion as the reference. OBT and WSF Tracker were evaluated at 10\,m and 100\,m, while TEMPO and GHSL were evaluated at 100\,m only. No product was disaggregated below its native spatial support. This procedure preserves representation-specific classification while establishing common spatial support and built-up-area criteria for comparison.

\medskip
\noindent\textbf{Detection and quantity agreement.}
Candidate and reference masks were compared cell by cell to obtain TP, FP, and FN counts, from which precision, recall, and F1 were calculated. Relative built-up-area bias was assessed using Eq.~\ref{eq:area_bias}, with $A_{\mathrm{pred}}$ and $A_{\mathrm{ref}}$ representing candidate built-up area and reference building area at the corresponding evaluation scale.

To distinguish errors in total mapped quantity from spatial misallocation, raster disagreement was decomposed into quantity disagreement (QD) and allocation disagreement (AD) \citep{pontius2011}:

\begin{equation}
QD =
\frac{|FP-FN|}{N},
\label{eq:quantity_disagreement}
\end{equation}

where $QD$ measures quantity disagreement, $FP$ and $FN$ are false-positive and false-negative grid-cell counts, and $N$ is the total number of evaluated cells.

\begin{equation}
AD =
\frac{2\min(FP,FN)}{N},
\label{eq:allocation_disagreement}
\end{equation}

where $AD$ measures allocation disagreement, $\min(FP,FN)$ is the smaller error count, and $N$ is the total number of evaluated cells. Together,

\begin{equation}
QD+AD =
\frac{FP+FN}{N},
\label{eq:total_disagreement}
\end{equation}

where $QD+AD$ represents total disagreement, and $FP+FN$ is the total number of incorrectly classified cells. This decomposition distinguishes disagreement in mapped built-up quantity from disagreement in spatial allocation.

\subsection{Building-Count Assessment}
\label{subsec:building_count}

Vector building counts were obtained directly from mapped polygons. For raster products, which generally represent built-up area rather than individual structures, approximate building counts were derived as

\begin{equation}
n_{\mathrm{pred}}
=
\frac{A_{\mathrm{pred}}}{\bar{A}_{\mathrm{ref}}},
\label{eq:estimated_count}
\end{equation}

where $n_{\mathrm{pred}}$ is the estimated building count, $A_{\mathrm{pred}}$ is the candidate built-up area, and $\bar{A}_{\mathrm{ref}}$ is the mean reference-building area within the same evaluation unit.

Count agreement was summarized using Eq.~\ref{eq:count_ratio}. Because raster-derived counts depend on reference building size, they were treated as exploratory quantity diagnostics rather than direct estimates of individual-building detection.

\subsection{Stratified Evaluation and Diagnostic Experiments}
\label{subsec:robustness_checks}

To identify systematic variation concealed by aggregate performance, the benchmark incorporated stratified analyses of reference provenance and settlement morphology, alongside targeted small-building and temporal-alignment diagnostics.

\noindent\textbf{Reference provenance.}
Performance was stratified by reference source---HOTOSM, SpaceNet~7, and other institutional or governmental datasets---to assess variation associated with mapping practices, geographic coverage, and potential shared data provenance.

\medskip
\noindent\textbf{Settlement morphology.}
Tiles were grouped into fixed classes of reference building density and mean building size, consistently applied across products and study areas. Relationships between these characteristics and tile-level F1 were assessed using Spearman's rank correlation coefficient, $\rho$, a non-parametric measure of monotonic association. Positive or negative $\rho$ indicates increasing or decreasing performance with the corresponding settlement characteristic.

\medskip
\noindent\textbf{Small-building diagnostic.}
Vector matching was repeated using the original matching thresholds and $20\,\mathrm{m}^2$ minimum-area filter, retaining the areas of TP, FP, and FN footprints. Buildings were grouped into six area classes: 20--40, 40--60, 60--80, 80--120, 120--180, and $>180\,\mathrm{m}^2$. Class-specific FP and FN proportions were calculated relative to candidate and reference buildings, respectively, to identify size-dependent detection errors.

\medskip
\noindent\textbf{Temporal-alignment diagnostic.}
Temporal mismatch was examined using WSF Tracker, whose six-monthly observations span the reference-data periods. For study areas with known reference-imagery dates, WSF Tracker was reevaluated using the closest available observation. The effect of temporal alignment was quantified as

\begin{equation}
\Delta F_1 =
F_{1,\mathrm{aligned}} -
F_{1,\mathrm{baseline}},
\label{eq:temporal_f1_change}
\end{equation}

where $\Delta F_1$ is the change in city-level F1, $F_{1,\mathrm{aligned}}$ is the score obtained using the observation closest to the reference date, and $F_{1,\mathrm{baseline}}$ is the score from the most recent WSF Tracker observation used in the main benchmark. Positive values indicate improved agreement following temporal alignment.

\section{Benchmark Performance and Contextual Variation}
\label{sec:results}

\subsection{Detection and Geometric Agreement}
\label{subsec:overall_performance}

Table~\ref{tab:overall_metrics} summarizes performance across the seven products, reported as medians of city-level metrics. Product coverage varied: Overture Maps and GBA covered 135 study areas, 3D-GloBFP 131, OBT 70, and WSF Tracker, TEMPO, and GHSL 134 each. Vector and raster results are evaluated within their respective representations, with raster comparisons further distinguished by evaluation resolution.

\begin{figure}[!t]
    \centering
    \includegraphics[width=\linewidth]{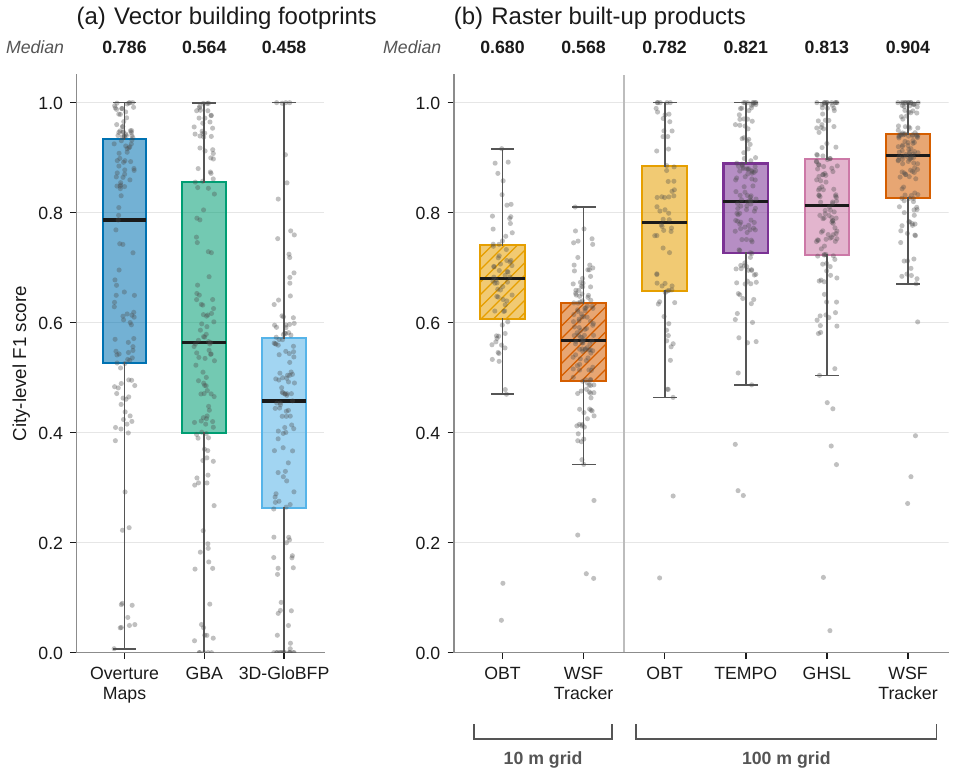}
    \caption{City-level F1 distributions for (a) vector building-footprint and (b) raster building and settlement products. Points represent study areas; boxes indicate interquartile ranges, horizontal lines medians, and whiskers extend to 1.5$\times$IQR. Raster products are grouped by 10\,m and 100\,m evaluation resolutions.}
    \label{fig:overall_detection_performance}
\end{figure}

\noindent\textbf{Vector building footprints.}
Overture Maps achieved the highest object-detection accuracy (median F1 = 0.786), followed by GBA (0.564) and 3D-GloBFP (0.458) (Figure~\ref{fig:overall_detection_performance}a). The same ranking was observed for matched-footprint geometry (Table~\ref{tab:vector_geometric_metrics}), with Overture achieving median matched-pair IoU and boundary F-scores of 0.965 and 0.974, respectively. Despite substantial variation across study areas, Overture consistently exhibited the strongest overall vector performance.

\noindent\textbf{Raster building and settlement products.}
Performance rankings depended on evaluation resolution (Figure~\ref{fig:overall_detection_performance}b). At 10\,m, OBT achieved a higher median F1 (0.680) than WSF Tracker (0.568). At 100\,m, WSF Tracker led (0.904), followed by TEMPO (0.821), GHSL (0.813), and OBT (0.782). Higher F1 at 100\,m reflects greater agreement in broader settlement occurrence following spatial aggregation, whereas 10\,m evaluation remains more sensitive to local positional and delineation differences. Scores across these scales therefore represent different levels of spatial agreement and are not directly interchangeable.

\begin{table}[!th]
\centering
\caption{Geometric agreement of matched vector building footprints. Higher values indicate better agreement.}
\label{tab:vector_geometric_metrics}

\renewcommand{\arraystretch}{1.15}
\setlength{\tabcolsep}{5pt}
\small

\begin{tabular}{lcccc}
\toprule
& \multicolumn{2}{c}{\textbf{Matched IoU}} &
\multicolumn{2}{c}{\textbf{Boundary F}} \\
\cmidrule(lr){2-3}
\cmidrule(lr){4-5}
\textbf{Dataset}
& \textbf{Median}
& \textbf{Mean}
& \textbf{Median}
& \textbf{Mean} \\
\midrule
Overture Maps & 0.965 & 0.782 & 0.974 & 0.754 \\
GBA           & 0.625 & 0.686 & 0.712 & 0.685 \\
3D-GloBFP     & 0.582 & 0.594 & 0.609 & 0.617 \\
\bottomrule
\end{tabular}
\end{table}

\begin{table*}[!t]
\centering
\caption{Median performance metrics for vector and raster products. Bold and underlined values indicate the best and second-best performance within each evaluation group, respectively. Higher precision, recall, and F1 are better; area bias and count ratio are ranked by proximity to zero and one, respectively.}
\label{tab:overall_metrics}

\renewcommand{\arraystretch}{1.15}
\setlength{\tabcolsep}{6pt}
\small

\begin{tabular}{lcccccc}
\toprule
\textbf{Dataset} &
\textbf{Precision} &
\textbf{Recall} &
\textbf{F1} &
\textbf{Relative area bias} &
\textbf{Count ratio} &
\textbf{$n$} \\
\midrule

\multicolumn{7}{l}{\textit{Vector building footprints}} \\

Overture Maps
& \textbf{0.720}
& \textbf{0.926}
& \textbf{0.786}
& \underline{0.093}
& \underline{1.163}
& 135 \\

Global Building Atlas (GBA)
& \underline{0.533}
& \underline{0.721}
& \underline{0.564}
& 0.124
& 1.167
& 135 \\

3D-GloBFP
& 0.462
& 0.449
& 0.458
& \textbf{$-0.045$}
& \textbf{0.904}
& 131 \\

\midrule
\multicolumn{7}{l}{\textit{Raster products evaluated at 10\,m}} \\

Google Open Buildings Temporal (OBT)
& \textbf{0.568}
& \textbf{0.878}
& \textbf{0.680}
& \textbf{0.123}
& \textbf{1.112}
& 70 \\

WSF Tracker
& \underline{0.485}
& \underline{0.796}
& \underline{0.568}
& \underline{1.719}
& \underline{2.612}
& 134 \\

\midrule
\multicolumn{7}{l}{\textit{Raster products evaluated at 100\,m}} \\

Google Open Buildings Temporal (OBT)
& 0.651
& \textbf{0.995}
& 0.782
& \underline{0.165}
& \underline{1.096}
& 70 \\

Microsoft TEMPO
& 0.703
& \textbf{0.995}
& \underline{0.821}
& \textbf{0.146}
& \textbf{1.088}
& 134 \\

GHSL
& \underline{0.727}
& \underline{0.978}
& 0.813
& 0.544
& 1.479
& 134 \\

WSF Tracker
& \textbf{0.888}
& 0.974
& \textbf{0.904}
& 1.845
& 2.585
& 134 \\

\bottomrule
\end{tabular}

\vspace{1mm}
\begin{minipage}{0.97\textwidth}
\footnotesize
\textit{Note:} Results at 10\,m and 100\,m represent different spatial supports. Vector counts represent mapped polygons, whereas raster counts are area-based proxies derived using mean reference-building size. Positive and negative area bias indicate overestimation and underestimation relative to reference footprint area.
\end{minipage}
\end{table*}

\subsection{Aggregate Quantity Agreement}
\label{subsec:quantity_agreement}

\begin{figure*}[!t]
    \centering
    \includegraphics[width=\textwidth]{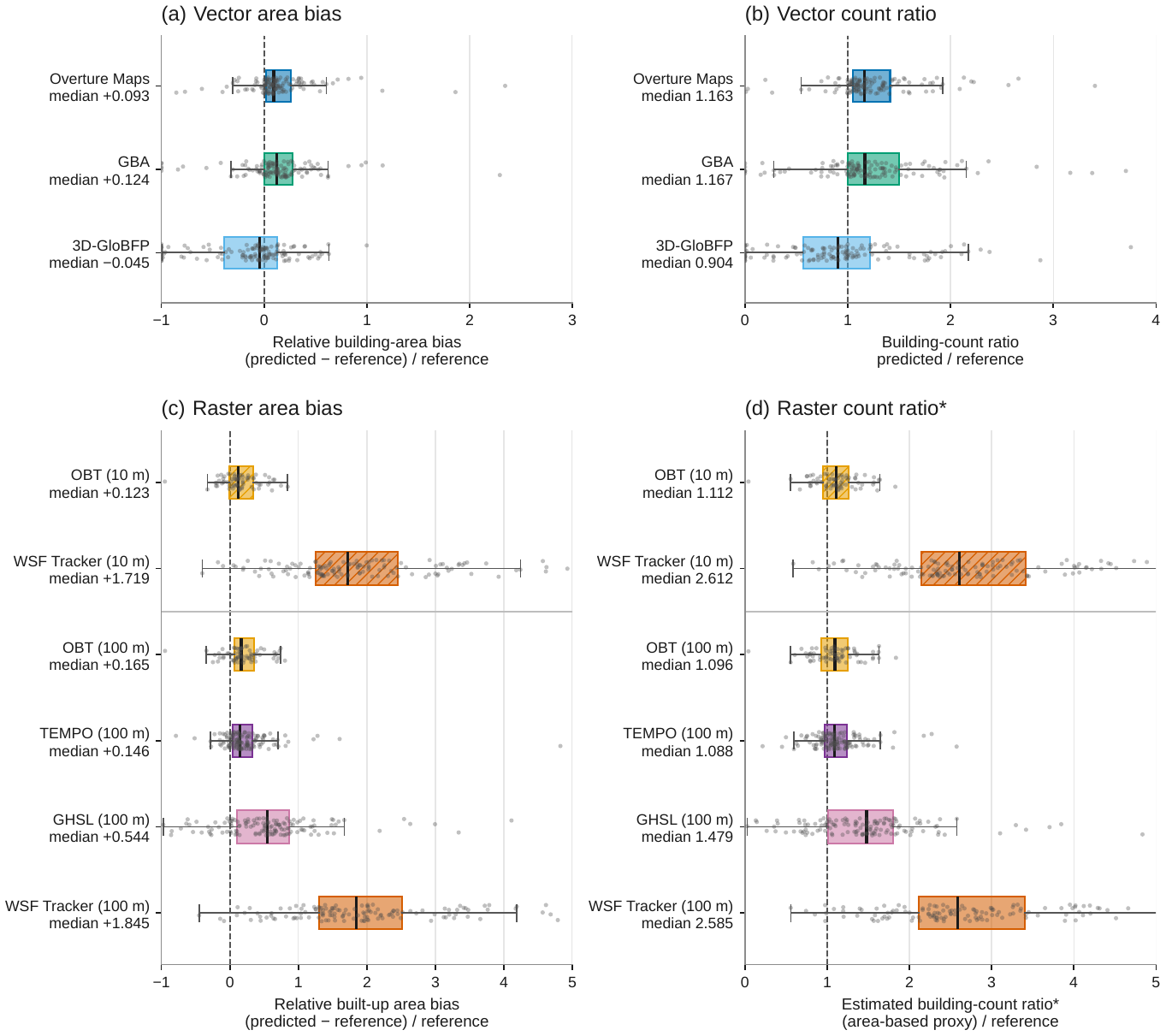}
    \caption{Aggregate quantity agreement for vector and raster products. (a, b) Relative building-area bias and building-count ratios for vector products; (c, d) relative built-up-area bias and estimated count ratios for raster products. Dashed lines indicate perfect agreement. Raster counts are area-based proxies rather than direct counts of individual buildings.}
    \label{fig:complementary_quantity_metrics}
\end{figure*}

Detection accuracy did not necessarily correspond to accurate estimates of aggregate building stock (Figure~\ref{fig:complementary_quantity_metrics}).

\noindent\textbf{Vector building area and counts.}
Overture and GBA modestly overestimated reference building quantities, with median relative area biases of 0.093 and 0.124 and count ratios of 1.163 and 1.167, respectively. Conversely, 3D-GloBFP achieved the closest aggregate agreement (area bias = $-0.045$; count ratio = 0.904), despite having the lowest object-detection F1 (0.458). Thus, accurate aggregate quantities did not necessarily imply correct detection or delineation of individual buildings.

\noindent\textbf{Raster built-up area and counts.}
At 10\,m, OBT showed substantially closer area agreement (bias = 0.123) than WSF Tracker (1.719). At 100\,m, TEMPO achieved the smallest area bias (0.146), followed by OBT (0.165), GHSL (0.544), and WSF Tracker (1.845). This contrast was particularly pronounced for WSF Tracker, which achieved the highest 100\,m detection F1 (0.904) but the greatest overestimation relative to reference building area. Its strong coarse-scale detection therefore reflects settlement occurrence rather than equivalence with individual building-footprint area. TEMPO and OBT, meanwhile, showed closer aggregate quantity agreement despite lower F1.

Raster-derived count ratios followed similar patterns but should be interpreted as exploratory area-based proxies rather than direct building counts. Quantity and allocation disagreement are reported separately in Supplementary Table~\ref{tab:supp_quantity_allocation}.

\subsection{Contextual Variation and Diagnostic Analyses}
\label{subsec:context_variation}

Product performance varied across reference sources, settlement characteristics, building sizes, and observation dates, revealing systematic differences not captured by overall benchmark scores.

\begin{figure*}[!t]
    \centering
    \includegraphics[width=\textwidth]{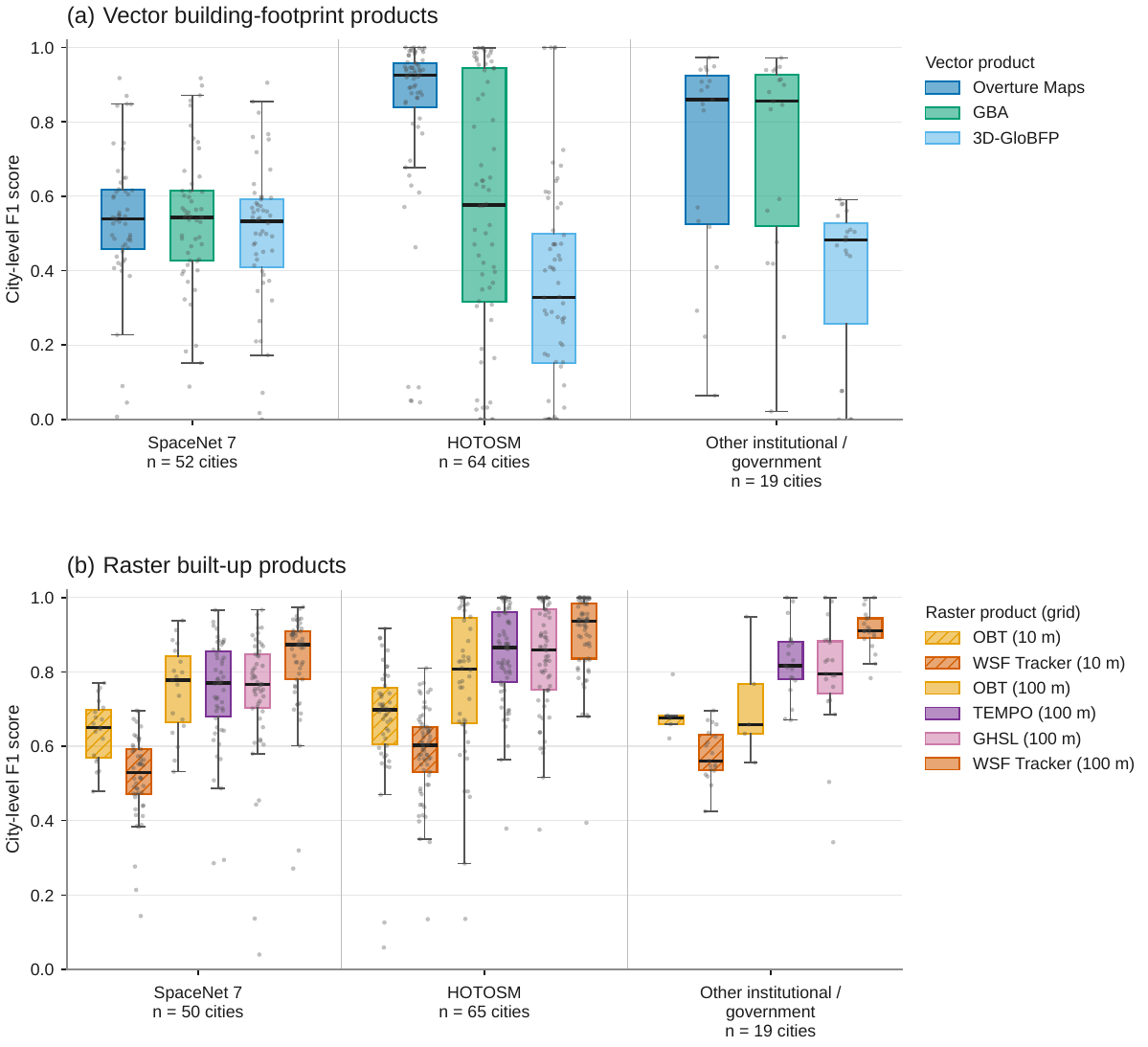}
    \caption{City-level F1 by reference-data source for (a) vector and (b) raster products. Boxplots show performance distributions across study areas, with group sample sizes indicated. Reference-source differences may also reflect geographic, temporal, and annotation characteristics.}
    \label{fig:f1_by_reference_source}
\end{figure*}

\noindent\textbf{Reference provenance.}
Vector accuracy varied substantially with reference-data source (Figure~\ref{fig:f1_by_reference_source}). Overture achieved higher F1 against HOTOSM and institutional references than against SpaceNet~7, whereas GBA exhibited greater within-group variability and 3D-GloBFP remained comparatively lower. Raster rankings were more consistent: at 100\,m, WSF Tracker maintained median F1 values of approximately 0.82, 0.89, and 0.86 against SpaceNet~7, HOTOSM, and other institutional references, respectively. These differences may reflect geographic and annotation characteristics as well as shared upstream data provenance, particularly between Overture and HOTOSM-derived references.

\begin{figure*}[!t]
    \centering
    \includegraphics[width=\textwidth]{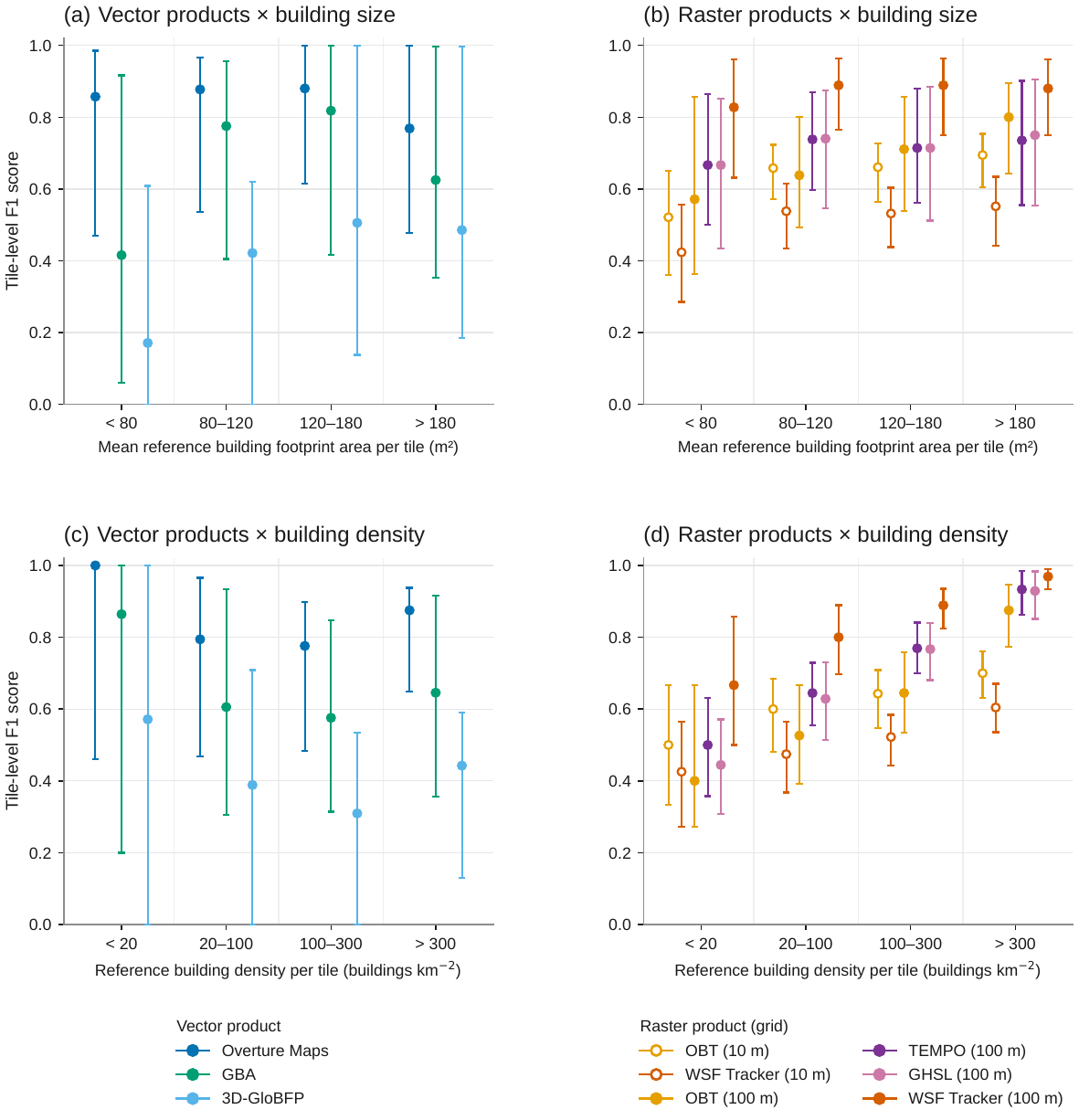}
    \caption{Tile-level F1 by mean reference building size (a, b) and building density (c, d) for vector (a, c) and raster (b, d) products. Markers indicate medians and vertical ranges the interquartile ranges. Open and filled raster markers denote 10\,m and 100\,m evaluations, respectively.}
    \label{fig:per_tile_density_size}
\end{figure*}

\noindent\textbf{Building density and size.}
Raster accuracy increased strongly with reference building density (Spearman's $\rho=0.58$--$0.75$, $p<0.001$), with the lowest F1 generally observed below 20 buildings\,km$^{-2}$ and the highest above 300 buildings\,km$^{-2}$ (Figure~\ref{fig:per_tile_density_size}). Vector performance showed weaker and less consistent density dependence. Overture remained comparatively stable, while GBA and 3D-GloBFP exhibited greater variation, particularly in sparse and intermediate-density settlements. Building-size effects were less pronounced overall, although GBA and especially 3D-GloBFP performed worse in tiles dominated by buildings smaller than 80\,m$^2$.

\noindent\textbf{Small-building errors.}
Vector false positives were disproportionately associated with smaller footprints. Median false-positive areas ranged from 57.7 to 71.6\,m$^2$, compared with 87.4--115.5\,m$^2$ for matched true positives. City-level count ratios also deviated more strongly from unity than area ratios, suggesting that overprediction frequently involved additional small polygons without proportional increases in mapped building area. This pattern was most pronounced for GBA and 3D-GloBFP and is consistent with footprint fragmentation or inclusion of additional small structures (Supplementary Figures~\ref{fig:supp_count_area}--\ref{fig:supp_small_buildings}).

\noindent\textbf{Temporal alignment.}
Matching WSF Tracker observations to reference-imagery dates improved city-level F1 by a mean of $\Delta F_1=0.060$ (median = 0.037; maximum = 0.351). Later observations generally increased recall but reduced precision, as newly developed settlements appeared as false positives against older reference data. These findings demonstrate that temporal mismatch can substantially affect apparent product accuracy, particularly in changing settlements (Supplementary Section~\ref{subsec:supp_temporal_alignment}).

\section{Discussion}
\label{sec:discussion}
\subsection{Representation and Spatial Scale Shape Product Accuracy}
The findings demonstrate that global building and settlement products cannot be treated as interchangeable representations of the built environment. Vector products delineate individual structures, whereas raster products represent building presence, built-up quantities, or broader settlement extent. Consequently, detection, geometric agreement, and aggregate quantities capture complementary dimensions of accuracy. This distinction is evident in the contrasting performance of Overture and 3D-GloBFP: the former achieved stronger object detection and geometric agreement, while the latter reproduced aggregate building quantities more closely despite weaker detection performance.

Spatial support further influences apparent accuracy. OBT performed best at 10\,m, whereas WSF Tracker achieved the highest F1 at 100\,m but substantially overestimated reference building area. Previous EO research has highlighted the influence of threshold selection and spatial support on built-up classification accuracy \citep{uhl2023builtup}. Our findings extend this perspective across vector and raster representations, demonstrating that spatial aggregation can improve detection scores without improving fine-scale building delineation. Moreover, quantity disagreement accounted for 83--99\% of total raster disagreement at 100\,m, reinforcing the importance of assessing mapped quantities independently of spatial detection.

\subsection{Settlement Morphology and EO Observability}
Settlement morphology emerged as an important correlate of product reliability. Raster F1 increased strongly with building density ($\rho=0.58$--$0.75$), whereas vector performance showed weaker density dependence but greater sensitivity to building size. Smaller candidate footprints were disproportionately associated with false positives, particularly for GBA and 3D-GloBFP.

These patterns are consistent with previous assessments identifying geographic and socioeconomic variation in building-product accuracy \citep{gevaert2024,lahrichi2026buildingarea}. Similarly, \citet{li2022crossgeonet} examined geographic transferability challenges in EO-based building extraction across label-scarce regions. Our findings complement these studies by linking performance variation to interpretable settlement characteristics, suggesting that geographic differences in accuracy partly reflect differences in building observability. This is especially relevant to sparse rural, peri-urban, and informal settlements, where small or closely spaced structures challenge reliable extraction. Morphology-stratified evaluation therefore provides a useful complement to geographic accuracy comparisons.

\subsection{Reference Provenance and Temporal Alignment}
Reference provenance and observation time introduce additional uncertainty into EO product validation. Overture's high agreement with some HOTOSM references may partly reflect shared OpenStreetMap sources rather than fully independent retrieval accuracy. Reference-source differences also covary with geographic, morphological, and annotation characteristics, limiting their interpretation as independent effects.

Temporal mismatch likewise affects apparent accuracy. Aligning WSF Tracker observations with reference-imagery dates increased mean city-level F1 by 0.060, indicating that some apparent errors may reflect genuine settlement change. Recent EO studies on building-footprint updating and multitemporal built-up monitoring emphasize the importance of repeated observations for tracking urban change \citep{zhou2026buildrenew,sapena2026builtup}. Our results demonstrate a complementary validation challenge: increasingly frequent EO observations require temporally corresponding reference data to distinguish mapping errors from actual landscape change. Reference independence and temporal alignment should therefore be considered integral to product evaluation.

\subsection{Product Suitability and Framework Extensibility}
The framework shifts emphasis from identifying a universally superior building dataset toward determining which products are appropriate for particular urban applications. Population mapping and exposure assessment may prioritize reliable aggregate building quantities, whereas infrastructure and morphological analyses require stronger geometric correspondence. Urban-change monitoring additionally depends on temporal consistency. Product suitability must therefore be assessed against the representation, spatial scale, and accuracy dimensions required by the intended application.

The framework also provides a transferable structure for evaluating emerging EO-derived and EO-integrated urban datasets. Recent developments in multimodal Sentinel-1/Sentinel-2 built-up extraction illustrate the increasing diversity of EO data sources and mapping approaches \citep{li2025builtup}. Such products can be incorporated through native-resolution classification and common-grid evaluation while retaining contextual and temporal diagnostics. Extensions to building-height or other attribute products would require corresponding reference measurements and attribute-specific metrics. This modular approach supports broader application of the framework without requiring heterogeneous products to share identical output representations.

\subsection{Scope and Interpretive Considerations}
\label{subsec:limitations}
Several considerations qualify the findings. First, temporal sensitivity was assessed only for WSF Tracker because comparable time-aligned observations were unavailable for all products. Second, shared upstream sources, particularly between Overture and HOTOSM references, may inflate apparent agreement. Third, SpaceNet~7 used an IoU matching threshold of 0.25 rather than 0.50, introducing differences in evaluation criteria across reference groups \citep{vanetten2021challenge}. Finally, the findings are limited to the sampled study areas and evaluated spatial supports (1\,km tiles and 10\,m or 100\,m raster grids). They provide comparative evidence of product reliability but do not establish suitability for cadastral or other individual-building applications requiring authoritative local verification.

\section{Conclusion}
\label{sec:conclusion}

This study introduces a representation-aware evaluation framework for benchmarking global building and settlement products across diverse geographic contexts. By integrating complementary detection, geometric, and quantity metrics with stratified and diagnostic analyses, the framework moves beyond aggregate rankings to identify where, when, and under what conditions different products are reliable.

Evaluation of seven products across 135 study areas revealed substantial differences in performance across representations, spatial scales, and settlement characteristics. Overture Maps achieved the highest vector building-detection accuracy (median F1 = 0.786), while 3D-GloBFP showed closer agreement in aggregate building quantities despite weaker detection. Among raster products, OBT performed best at 10\,m (F1 = 0.680), whereas WSF Tracker achieved the highest F1 at 100\,m (0.904) but substantially overestimated reference building area. Accuracy also varied with building density, size, reference provenance, and temporal alignment, demonstrating that strong detection performance does not necessarily imply accurate building quantities or consistent geographic reliability.

These findings underscore the importance of selecting building and settlement products according to their representation, evaluation scale, local settlement characteristics, and intended application. Global products provide valuable information where authoritative data are limited, but their suitability should be established through complementary accuracy measures and, where necessary, local validation rather than a single overall performance score.


\section*{Code and data availability}


Benchmark outputs and associated data products will be deposited in the World Bank Data Catalog upon publication.

All code used for data processing, benchmarking, evaluation, and analysis is publicly available on GitHub: \url{https://github.com/GFDRR/urban_validation}.


\section*{Acknowledgements}

This research was supported by the Gates Foundation through the World Bank's Global Facility for Disaster Reduction and Recovery (GFDRR) Digital Earth Partnership. The authors gratefully acknowledge the organizations and initiatives that provided reference and validation datasets, including the World Bank and SPARC-related initiatives, the Humanitarian OpenStreetMap Team (HOTOSM), the University of Edinburgh, and the SpaceNet~7 Multi-Temporal Urban Development Challenge. We also thank the developers and maintainers of the global building and settlement products evaluated in this study for making their datasets publicly accessible. Additional funding support was provided by the American Association for the Advancement of Science (AAAS) through the Roger Revelle Fellowship in Global Stewardship.

\section*{Declaration of Generative AI Use}
During the preparation of this manuscript, the authors used OpenAI's ChatGPT to assist with language refinement, editing, and improvements to the clarity and organization of the text. Anthropic's Claude (Opus 5.5) was used to generate TikZ-based diagrams with handwritten-style typography for Figures 1, 3 and 4. All AI-assisted content was reviewed, verified, and revised by the authors, who take full responsibility for the accuracy, interpretation, and final content of the manuscript. Generative AI was not used for data generation, processing, analysis, or interpretation of the study results.


\bibliography{reference}

@article{bettencourt2025infrastructure,
  author  = {Bettencourt, Lu{\'i}s M. A. and Marchio, Nicholas},
  title   = {Infrastructure Deficits and Informal Settlements in Sub-Saharan Africa},
  journal = {Nature},
  volume  = {645},
  number  = {8080},
  pages   = {399--406},
  year    = {2025},
  doi     = {10.1038/s41586-025-09465-2}
}

@article{boo2022,
  author  = {Boo, Gianluca and Darin, Edith and Leasure, Douglas R. and Dooley, Claire A. and Chamberlain, Heather R. and L{\'a}zar, Attila N. and Tschirhart, Kevin and Sinai, Cyrus and Hoff, Nicholas A. and Fuller, Trevon and Musene, Kevin and Batumbo, Arly and Rimoin, Anne W. and Tatem, Andrew J.},
  title   = {High-resolution population estimation using household survey data and building footprints},
  journal = {Nat. Commun.},
  volume  = {13},
  number  = {1},
  pages   = {1330},
  year    = {2022},
  doi     = {10.1038/s41467-022-29094-x}
}

@article{chamberlain2024africa,
  author  = {Chamberlain, Heather R. and Darin, Edith and Adewole, Wole Ademola and Jochem, Warren C. and Lazar, Attila N. and Tatem, Andrew J.},
  title   = {Building Footprint Data for Countries in Africa: To What Extent Are Existing Data Products Comparable?},
  journal = {Comput. Environ. Urban Syst.},
  volume  = {110},
  pages   = {102104},
  year    = {2024},
  doi     = {10.1016/j.compenvurbsys.2024.102104}
}

@article{che20243dglobfp,
  author  = {Che, Yangzi and Li, Xuecao and Liu, Xiaoping and Wang, Yuhao and Liao, Weilin and Zheng, Xianwei and Zhang, Xucai and Xu, Xiaocong and Shi, Qian and Zhu, Jiajun and Zhang, Honghui and Yuan, Hua and Dai, Yongjiu},
  title   = {{3D-GloBFP}: The First Global Three-Dimensional Building Footprint Dataset},
  journal = {Earth Syst. Sci. Data},
  volume  = {16},
  pages   = {5357--5376},
  year    = {2024},
  doi     = {10.5194/essd-16-5357-2024}
}

@misc{dlr2026wsftracker,
  author       = {{German Aerospace Center}},
  title        = {World Settlement Footprint Tracker},
  year         = {2026},
  howpublished = {German Aerospace Center, Earth Observation Center, AI4SmartCities CCN3 project documentation},
  note         = {Accessed: 27 July 2026},
  url          = {https://www.dlr.de/en/eoc/research-transfer/projects-missions/ai4smartcities-ccn3}
}

@article{esch2022wsf3d,
  author  = {Esch, Thomas and Brzoska, Elisabeth and Dech, Stefan and Leutner, Benjamin and Palacios-Lopez, Daniela and Metz-Marconcini, Annekatrin and Marconcini, Mattia and Roth, Achim and Zeidler, Julian},
  title   = {{World Settlement Footprint 3D}---a first three-dimensional survey of the global building stock},
  journal = {Remote Sens. Environ.},
  volume  = {270},
  pages   = {112877},
  year    = {2022},
  doi     = {10.1016/j.rse.2021.112877}
}

@misc{firth2020hot,
  author = {Firth, Rebecca},
  title  = {{HOT} Community Strategy 2020},
  year   = {2020},
  note   = {Humanitarian OpenStreetMap Team community strategy}
}

@article{gabrielli2026paneuropean,
  author  = {Gabrielli, Lorenzo and Sulis, Patrizia and Thabit, Sara and Minghini, Marco},
  title   = {Towards a Comparison of the Semantic Information of Pan-{European} Open Building Data},
  journal = {ISPRS Int. J. Geo-Inf.},
  volume  = {15},
  number  = {6},
  pages   = {252},
  year    = {2026},
  doi     = {10.3390/ijgi15060252}
}

@article{gevaert2024,
  author  = {Gevaert, Caroline M. and Buunk, Thomas and van den Homberg, Marc J. C.},
  title   = {Auditing Geospatial Datasets for Biases: Using Global Building Datasets for Disaster Risk Management},
  journal = {IEEE J. Sel. Top. Appl. Earth Obs. Remote Sens.},
  volume  = {17},
  pages   = {12579--12590},
  year    = {2024},
  doi     = {10.1109/JSTARS.2024.3422503}
}

@misc{glazer2025tempo,
  author        = {Glazer, Tammy and Hacheme, Gilles Q. and Zaytar, Akram and Marotti, Luana and Michaels, Amy and Tadesse, Girmaw Abebe and White, Kevin and Dodhia, Rahul and Zolli, Andrew and Becker-Reshef, Inbal and Lavista Ferres, Juan M. and Robinson, Caleb},
  title         = {{TEMPO}: Global Temporal Building Density and Height Estimation from Satellite Imagery},
  year          = {2025},
  howpublished  = {arXiv preprint, arXiv:2511.12104},
  doi           = {10.48550/arXiv.2511.12104},
  url           = {https://arxiv.org/abs/2511.12104},
  eprint        = {2511.12104},
  archivePrefix = {arXiv},
  primaryClass  = {cs.CV}
}

@article{herfort2021evolution,
  author  = {Herfort, Benjamin and Lautenbach, Sven and Porto de Albuquerque, Jo{\~a}o and Anderson, Jennings and Zipf, Alexander},
  title   = {The Evolution of Humanitarian Mapping within the {OpenStreetMap} Community},
  journal = {Sci. Rep.},
  volume  = {11},
  number  = {1},
  pages   = {3037},
  year    = {2021},
  doi     = {10.1038/s41598-021-82404-z}
}

@misc{lahrichi2026buildingarea,
  author        = {Lahrichi, Saad and Allabadi, {Doa'a} and Bradbury, Kyle and Malof, Jordan},
  title         = {Global Building Area Estimation Products: How Accurate Are They?},
  year          = {2026},
  howpublished  = {arXiv preprint, arXiv:2607.19766},
  doi           = {10.48550/arXiv.2607.19766},
  url           = {https://arxiv.org/abs/2607.19766},
  eprint        = {2607.19766},
  archivePrefix = {arXiv},
  primaryClass  = {cs.CV}
}

@misc{linuxfoundation2022overture,
  author = {{Linux Foundation}},
  title  = {Joint Development Foundation Announces {Overture Maps Foundation} to Build Interoperable Open Map Data},
  year   = {2022},
  note   = {Press release, 15 December 2022},
  url    = {https://www.linuxfoundation.org/press/linux-foundation-announces-overture-maps-foundation}
}

@article{liu2020ghsl,
  author  = {Liu, Feng and Wang, Shuai and Xu, Yi and Ying, Qing and Yang, Fukun and Qin, Yuchu},
  title   = {Accuracy Assessment of Global Human Settlement Layer ({GHSL}) Built-Up Products over China},
  journal = {PLoS ONE},
  volume  = {15},
  number  = {5},
  pages   = {e0233164},
  year    = {2020},
  doi     = {10.1371/journal.pone.0233164}
}

@article{lloyd2019,
  author  = {Lloyd, Christopher T. and Chamberlain, Heather and Kerr, David and Yetman, Greg and Pistolesi, Linda and Stevens, Forrest R. and Gaughan, Andrea E. and Nieves, Jeremiah J. and Hornby, Graeme and MacManus, Kytt and Sinha, Parmanand and Bondarenko, Maksym and Sorichetta, Alessandro and Tatem, Andrew J.},
  title   = {Global spatio-temporally harmonised datasets for producing high-resolution gridded population distribution datasets},
  journal = {Big Earth Data},
  volume  = {3},
  number  = {2},
  pages   = {108--139},
  year    = {2019},
  doi     = {10.1080/20964471.2019.1625151}
}

@article{marconcini2020wsf,
  author  = {Marconcini, Mattia and Metz-Marconcini, Annekatrin and {\"U}reyen, Soner and Palacios-Lopez, Daniela and Hanke, Wiebke and Bachofer, Felix and Zeidler, Julian and Esch, Thomas and Gorelick, Noel and Kakarla, Ashwin and Paganini, Marc and Strano, Emanuele},
  title   = {Outlining Where Humans Live: The World Settlement Footprint 2015},
  journal = {Sci. Data},
  volume  = {7},
  number  = {1},
  pages   = {242},
  year    = {2020},
  doi     = {10.1038/s41597-020-00580-5}
}

@article{marconcini2021wsfsuite,
  author  = {Marconcini, Mattia and Metz-Marconcini, Annekatrin and Esch, Thomas and Gorelick, Noel},
  title   = {Understanding Current Trends in Global Urbanisation---the {World Settlement Footprint} Suite},
  journal = {GI\_Forum},
  volume  = {9},
  number  = {1},
  pages   = {33--38},
  year    = {2021},
  doi     = {10.1553/giscience2021_01_s33}
}

@techreport{oecd2025buildingdata,
  author      = {{OECD}},
  title       = {Building Data Together: Proceedings of an {OECD} Geospatial Lab Workshop},
  year        = {2025},
  institution = {OECD Publishing},
  address     = {Paris},
  note        = {71 pp.},
  doi         = {10.1787/725f7274-en}
}

@misc{okyere2025obd,
  author        = {Okyere, Franz and Lu, Meng and Brunn, Ansgar},
  title         = {Evaluating the Quality of Open Building Datasets for Mapping Urban Inequality: A Comparative Analysis Across Five Cities},
  year          = {2025},
  howpublished  = {arXiv preprint, arXiv:2508.12872},
  doi           = {10.48550/arXiv.2508.12872},
  url           = {https://arxiv.org/abs/2508.12872},
  eprint        = {2508.12872},
  archivePrefix = {arXiv},
  primaryClass  = {cs.CV}
}

@article{oostwegel2025openbuildingmap,
  author  = {Oostwegel, Laurens J. N. and Schorlemmer, Danijel and Gu{\'e}guen, Philippe},
  title   = {From Footprints to Functions: A Comprehensive Global and Semantic Building Footprint Dataset},
  journal = {Sci. Data},
  volume  = {12},
  number  = {1},
  pages   = {1699},
  year    = {2025},
  doi     = {10.1038/s41597-025-06132-z}
}

@misc{overture2026buildings,
  author       = {{Overture Maps Foundation}},
  title        = {Buildings Guide},
  year         = {2026},
  howpublished = {Overture Maps documentation},
  note         = {Accessed: 27 July 2026},
  url          = {https://docs.overturemaps.org/guides/buildings/}
}

@misc{pesaresi2023ghsbuilts,
  author       = {Pesaresi, Martino and Politis, Panagiotis},
  title        = {{GHS-BUILT-S R2023A}---{GHS} built-up surface grid, derived from {Sentinel-2} composite and {Landsat}, multitemporal (1975--2030)},
  year         = {2023},
  howpublished = {European Commission, Joint Research Centre (dataset)},
  doi          = {10.2905/9F06F36F-4B11-47EC-ABB0-4F8B7B1D72EA}
}

@article{pesaresi2024ghsl,
  author  = {Pesaresi, Martino and Schiavina, Marcello and Politis, Panagiotis and Freire, Sergio and Krasnod{\k{e}}bska, Katarzyna and Uhl, Johannes H. and Carioli, Alessandra and Corbane, Christina and Dijkstra, Lewis and Florio, Pietro and Friedrich, Hannah K. and Gao, Jing and Leyk, Stefan and Lu, Linlin and Maffenini, Luca and Mari-Rivero, Ines and Melchiorri, Michele and Syrris, Vasileios and Van Den Hoek, Jamon and Kemper, Thomas},
  title   = {Advances on the Global Human Settlement Layer by Joint Assessment of Earth Observation and Population Survey Data},
  journal = {Int. J. Digit. Earth},
  volume  = {17},
  number  = {1},
  pages   = {2390454},
  year    = {2024},
  doi     = {10.1080/17538947.2024.2390454}
}

@article{pontius2011,
  author  = {Pontius, Jr., Robert Gilmore and Millones, Marco},
  title   = {Death to {Kappa}: Birth of Quantity Disagreement and Allocation Disagreement for Accuracy Assessment},
  journal = {Int. J. Remote Sens.},
  volume  = {32},
  number  = {15},
  pages   = {4407--4429},
  year    = {2011},
  doi     = {10.1080/01431161.2011.552923}
}

@misc{sirko2021continental,
  author        = {Sirko, Wojciech and Kashubin, Sergii and Ritter, Marvin and Annkah, Abigail and Bouchareb, Yasser Salah Eddine and Dauphin, Yann and Keysers, Daniel and Neumann, Maxim and Cisse, Moustapha and Quinn, John},
  title         = {Continental-Scale Building Detection from High-Resolution Satellite Imagery},
  year          = {2021},
  howpublished  = {arXiv preprint, arXiv:2107.12283},
  doi           = {10.48550/arXiv.2107.12283},
  url           = {https://arxiv.org/abs/2107.12283},
  eprint        = {2107.12283},
  archivePrefix = {arXiv},
  primaryClass  = {cs.CV}
}

@misc{sirko2023sentinel2,
  author        = {Sirko, Wojciech and Brempong, Emmanuel Asiedu and Marcos, Juliana T. C. and Annkah, Abigail and Korme, Abel and Hassen, Mohammed Alewi and Sapkota, Krishna and Shekel, Tomer and Diack, Abdoulaye and Nevo, Sella and Hickey, Jason and Quinn, John},
  title         = {High-Resolution Building and Road Detection from {Sentinel-2}},
  year          = {2023},
  howpublished  = {arXiv preprint, arXiv:2310.11622},
  doi           = {10.48550/arXiv.2310.11622},
  url           = {https://arxiv.org/abs/2310.11622},
  eprint        = {2310.11622},
  archivePrefix = {arXiv},
  primaryClass  = {cs.CV}
}

@techreport{unhabitat2022world,
  author      = {{UN-Habitat}},
  title       = {World Cities Report 2022: Envisaging the Future of Cities},
  year        = {2022},
  institution = {United Nations Human Settlements Programme},
  address     = {Nairobi},
  note        = {422 pp.},
  url         = {https://unhabitat.org/world-cities-report-2022-envisaging-the-future-of-cities}
}

@misc{vanetten2021challenge,
  author        = {Van Etten, Adam and Hogan, Daniel},
  title         = {The {SpaceNet} Multi-Temporal Urban Development Challenge},
  year          = {2021},
  howpublished  = {arXiv preprint, arXiv:2102.11958},
  doi           = {10.48550/arXiv.2102.11958},
  url           = {https://arxiv.org/abs/2102.11958},
  eprint        = {2102.11958},
  archivePrefix = {arXiv}
}

@inproceedings{vanetten2021dataset,
  author    = {Van Etten, Adam and Hogan, Daniel and Manso, Jesus Martinez and Shermeyer, Jacob and Weir, Nicholas and Lewis, Ryan},
  title     = {The Multi-Temporal Urban Development {SpaceNet} Dataset},
  booktitle = {Proc. IEEE/CVF Conf. Comput. Vis. Pattern Recognit. (CVPR)},
  pages     = {6398--6407},
  year      = {2021},
  doi       = {10.1109/CVPR46437.2021.00633}
}

@misc{vida2023combined,
  author       = {{VIDA}},
  title        = {Google--Microsoft Open Buildings---Combined by {VIDA}},
  year         = {2023},
  howpublished = {Source Cooperative (dataset)},
  note         = {Published 15 September 2023},
  url          = {https://source.coop/vida/google-microsoft-open-buildings}
}

@article{yepes2023,
  author  = {Yepes-Estrada, Catalina and Calderon, Alejandro and Costa, Catarina and Crowley, Helen and Dabbeek, Jamal and Hoyos, Maria Camila and Martins, Luis and Paul, Nicole and Rao, Anirudh and Silva, Vitor},
  title   = {Global Building Exposure Model for Earthquake Risk Assessment},
  journal = {Earthq. Spectra},
  volume  = {39},
  number  = {4},
  pages   = {2212--2235},
  year    = {2023},
  doi     = {10.1177/87552930231194048}
}

@article{zhu2025gba,
  author  = {Zhu, Xiao Xiang and Chen, Sining and Zhang, Fahong and Shi, Yilei and Wang, Yuanyuan},
  title   = {{GlobalBuildingAtlas}: An Open Global and Complete Dataset of Building Polygons, Heights and {LoD1} 3D Models},
  journal = {Earth Syst. Sci. Data},
  volume  = {17},
  number  = {12},
  pages   = {6647--6668},
  year    = {2025},
  doi     = {10.5194/essd-17-6647-2025}
}

@article{GUNASEKERA2015594,
  author  = {Gunasekera, Rashmin and Ishizawa, Oscar and Aubrecht, Christoph and Blankespoor, Brian and Murray, Siobhan and Pomonis, Antonios and Daniell, James},
  title   = {Developing an Adaptive Global Exposure Model to Support the Generation of Country Disaster Risk Profiles},
  journal = {Earth-Sci. Rev.},
  volume  = {150},
  pages   = {594--608},
  year    = {2015},
  doi     = {10.1016/j.earscirev.2015.08.012}
}

@article{uhl2023builtup,
  author  = {Uhl, Johannes H. and Leyk, Stefan},
  title   = {Spatially Explicit Accuracy Assessment of Deep Learning-Based, Fine-Resolution Built-Up Land Data in the United States},
  journal = {Int. J. Appl. Earth Obs. Geoinf.},
  volume  = {123},
  pages   = {103469},
  year    = {2023},
  doi     = {10.1016/j.jag.2023.103469}
}

@article{li2022crossgeonet,
  author  = {Li, Qingyu and Mou, Lichao and Hua, Yuansheng and Shi, Yilei and Zhu, Xiao Xiang},
  title   = {{CrossGeoNet}: A Framework for Building Footprint Generation of Label-Scarce Geographical Regions},
  journal = {Int. J. Appl. Earth Obs. Geoinf.},
  volume  = {111},
  pages   = {102824},
  year    = {2022},
  doi     = {10.1016/j.jag.2022.102824}
}

@article{zhou2026buildrenew,
  author  = {Zhou, Wen and Wu, Chen and Du, Bo and Zhang, Liangpei},
  title   = {{SAM}-Powered Building Footprint Updating for Various Cities: Sparse Labels Meet Historical Data Repurposing in Urban Monitoring},
  journal = {Int. J. Appl. Earth Obs. Geoinf.},
  volume  = {146},
  pages   = {105062},
  year    = {2026},
  doi     = {10.1016/j.jag.2025.105062}
}

@article{sapena2026builtup,
  author  = {Sapena, Marta and Mast, Johannes and Schoepfer, Elisabeth and Taubenb{\"o}ck, Hannes},
  title   = {Do-It-Yourself Built-Up Mapping Tool: A Practical Cloud-Based Solution Using {Sentinel} Imagery for Mapping Urban Expansion in Africa},
  journal = {Int. J. Appl. Earth Obs. Geoinf.},
  volume  = {146},
  pages   = {105153},
  year    = {2026},
  doi     = {10.1016/j.jag.2026.105153}
}

@article{li2025builtup,
  author  = {Li, Yu and Matgen, Patrick and Chini, Marco},
  title   = {Extraction of Built-Up Areas Using {Sentinel-1} and {Sentinel-2} Data with Automated Training Data Sampling and Label Noise Robust Cross-Fusion Neural Networks},
  journal = {Int. J. Appl. Earth Obs. Geoinf.},
  volume  = {139},
  pages   = {104524},
  year    = {2025},
  doi     = {10.1016/j.jag.2025.104524}
}
\bibliographystyle{icml2025}


\clearpage
\appendix

\section{Supplementary Materials}
\label{sec:supp_diagnostics}

\subsection{Related Work}
\label{sec:supp_related_work}

\noindent\textbf{Vector building-footprint evaluations.}
\citet{chamberlain2024africa} compared Google Open Buildings, Ecopia, Microsoft Building Footprints, and OpenStreetMap (OSM) across Africa, reporting pairwise Jaccard coefficients generally between 0.40 and 0.60. These measures characterize inter-product agreement rather than accuracy against independent references. Similarly, \citet{okyere2025obd} reported substantial geographic variation in Microsoft--OSM geometric agreement across five cities, with mean IoU ranging from 0.84 in Houston to 0.46 in Caracas. Producer-led evaluations have focused more directly on instance detection. \citet{sirko2021continental} reported a mean average precision of 0.657 at an IoU threshold of 0.50 for Google Open Buildings across Africa and South Asia. \citet{zhu2025gba} compared several footprint products across 28 cities on five continents, reporting Asian $\mathrm{AP}_{50}$ scores of 6.5, 8.7, and 27.9 for Microsoft, 3D-GloBFP, and GBA, respectively. Africa was excluded because suitable authoritative references were unavailable. These studies differ in their reference independence, sampling strategies, and evaluation metrics.

\medskip
\noindent\textbf{Raster settlement-product evaluations.}
WSF2015 was validated using approximately 900,000 reference samples across 50 global tiles, achieving an average Kappa coefficient of 0.6885 \citep{marconcini2020wsf}. For GHSL R2023, \citet{pesaresi2024ghsl} reported an IoU of approximately 0.92 for binary built-up classification at 10\,m. An earlier evaluation across 20 Chinese cities reported an $R^2$ of 0.82 for GHSL built-up estimates aggregated to 1\,km, alongside density-dependent estimation biases \citep{liu2020ghsl}. These statistics describe different evaluation targets and spatial supports and should not be interpreted as directly comparable accuracy scores.

\medskip
\noindent\textbf{Broader quality and contextual assessments.}
Recent evaluations have incorporated additional dimensions of building-data quality. \citet{gabrielli2026paneuropean} compared six European building datasets, emphasizing semantic-attribute availability and completeness, while \citet{oecd2025buildingdata} examined data quality, usability, and governance, with quantitative comparisons centered on building counts and areas. \citet{lahrichi2026buildingarea} evaluated GHSL, TEMPO, GBA, and Overture against manually labelled ORBITaL-Net data, reporting differences across evaluation grids, geographic regions, settlement density, and national income groups. GBA or TEMPO generally performed best depending on the evaluation criterion. Similarly, \citet{gevaert2024} examined socioeconomic and settlement-related biases across four vector building datasets, although the analysis was restricted to two countries.

Shared upstream data can complicate independent validation. For example, OpenBuildingMap integrates OSM, Google, and Microsoft footprints \citep{oostwegel2025openbuildingmap}, while VIDA combines Google and Microsoft building products \citep{vida2023combined}. Such integrations can improve data coverage but may introduce dependencies between candidate and reference datasets. Consequently, high agreement should be interpreted in relation to dataset provenance, particularly when community-contributed mapping sources are used in both products and reference benchmarks.

\subsection{Raster Harmonization and Binarization}
\label{app:raster_binarisation}

Raster evaluation followed three stages: reference rasterization, native-resolution candidate classification, and aggregation to common evaluation grids.

\noindent\textbf{Reference rasterization.}
Reference footprints were rasterized to fractional built-up coverage using

\begin{equation}
f_{\mathrm{ref}}(q)
=
\frac{A_{\mathrm{ref}}(q)}{A_{\mathrm{eval}}},
\label{eq:ref_fraction}
\end{equation}

where $f_{\mathrm{ref}}(q)$ is the built-up fraction of evaluation cell $q$, $A_{\mathrm{ref}}(q)$ is its reference building-footprint area, and $A_{\mathrm{eval}}$ is the evaluation-cell area. Oversampling and block averaging preserved sub-pixel coverage. Binary reference labels were obtained as

\begin{equation}
y_{\mathrm{ref}}(q)
=
\mathbb{1}
\left[
f_{\mathrm{ref}}(q)
\geq \tau_{\mathrm{eval}}
\right],
\qquad
\tau_{\mathrm{eval}}
=
\min\left(1,\frac{M_{\mathrm{ref}}}{A_{\mathrm{eval}}}\right),
\label{eq:ref_binary}
\end{equation}

where $y_{\mathrm{ref}}(q)$ is the binary reference label, $\mathbb{1}[\cdot]$ is the indicator function, $\tau_{\mathrm{eval}}$ is the evaluation threshold, and $M_{\mathrm{ref}}=20\,\mathrm{m}^2$ is the minimum built-up area. This corresponds to fractional thresholds of 0.20 and 0.002 at 10\,m and 100\,m, respectively.

\medskip
\noindent\textbf{Native-resolution classification.}
Continuous candidate products were converted to native built-up area:

\begin{equation}
A_{\mathrm{pred,nat}}(p)
=
\begin{cases}
v(p)A_{\mathrm{nat}}(d),
& \text{fraction-based inputs},\\
\min[v(p),A_{\mathrm{nat}}(d)],
& \text{area-based inputs},
\end{cases}
\label{eq:native_area}
\end{equation}

where $A_{\mathrm{pred,nat}}(p)$ is the predicted built-up area in native pixel $p$, $v(p)$ is its input value, and $A_{\mathrm{nat}}(d)$ is the native pixel area of dataset $d$. Native classification was then defined as

\begin{equation}
y_{\mathrm{nat}}(p)
=
\mathbb{1}
\left[
A_{\mathrm{pred,nat}}(p)\geq M_d
\right],
\label{eq:native_binary}
\end{equation}

where $y_{\mathrm{nat}}(p)$ is the binary candidate label and $M_d$ is the product-specific minimum detectable built-up area. WSF Tracker was classified directly using its categorical settlement codes. Product thresholds are summarized in Table~\ref{tab:raster_thresholds}.

\begin{table}[!ht]
\centering
\caption{Native raster characteristics and classification thresholds.}
\label{tab:raster_thresholds}
\renewcommand{\arraystretch}{1.15}
\setlength{\tabcolsep}{4pt}
\small

\begin{tabular}{lcccc}
\toprule
\textbf{Product}
& \textbf{Grid (m)}
& \textbf{Area (m$^2$)}
& $M_d$ (m$^2$)
& $\tau_{\mathrm{nat}}$ \\
\midrule
OBT         & 4   & 16     & 4  & 0.25 \\
TEMPO       & 100 & 10,000 & 20 & 0.002 \\
GHSL        & 100 & 10,000 & 20 & 0.002 \\
WSF Tracker & 10  & 100    & -- & Categorical \\
\bottomrule
\end{tabular}
\end{table}

\noindent\textbf{Common-grid aggregation.}
Native binary masks were aggregated using area-weighted overlap:

\begin{equation}
f_{\mathrm{pred}}(q)
=
\frac{
\sum_{p\in P_q}a_{p,q}y_{\mathrm{nat}}(p)
}{
\sum_{p\in P_q}a_{p,q}
},
\label{eq:aggregation}
\end{equation}

where $f_{\mathrm{pred}}(q)$ is the predicted built-up fraction, $P_q$ contains valid native pixels intersecting evaluation cell $q$, and $a_{p,q}$ is their overlap area. Nodata pixels were excluded. The aggregated predictions were classified using the reference evaluation threshold:

\begin{equation}
y_{\mathrm{pred}}(q)
=
\mathbb{1}
\left[
f_{\mathrm{pred}}(q)\geq\tau_{\mathrm{eval}}
\right],
\label{eq:pred_binary}
\end{equation}

where $y_{\mathrm{pred}}(q)$ is the binary predicted label and $\tau_{\mathrm{eval}}$ is defined in Eq.~\ref{eq:ref_binary}.

For continuous products, built-up area was retained separately for area-based evaluation:

\begin{equation}
A_{\mathrm{pred}}(q)
=
A_{\mathrm{eval}}
\frac{
\sum_{p\in P_q}a_{p,q}
\frac{A_{\mathrm{pred,nat}}(p)}
{A_{\mathrm{nat}}(d)}
}{
\sum_{p\in P_q}a_{p,q}
}.
\label{eq:pred_area}
\end{equation}

Here, $A_{\mathrm{pred}}(q)$ is the aggregated predicted built-up area; the remaining variables are defined above. Thus, binary masks support detection assessment, while continuous area estimates retain sub-pixel information for quantity evaluation.

\subsection{Temporal Coverage and Benchmark Snapshots}
\label{app:temporal_coverage}

Table~\ref{tab:raster_dataset_dates} identifies the observation epochs used in the benchmark. For WSF Tracker, a common settlement cutoff was applied across study areas; temporally aligned observations were examined separately.

\begin{table}[!ht]
\centering
\caption{Temporal snapshots used for raster benchmarking.}
\label{tab:raster_dataset_dates}
\renewcommand{\arraystretch}{1.15}
\setlength{\tabcolsep}{4pt}
\small

\begin{tabular}{lccc}
\toprule
\textbf{Product}
& \textbf{Cadence}
& \textbf{Epoch used}
& \textbf{Grid} \\
\midrule
OBT         & Annual    & 2023    & 4\,m \\
TEMPO       & Quarterly & 2023 Q4 & 100\,m \\
WSF Tracker & Six-monthly & Jan 2026 & 10\,m \\
GHSL        & Epoch-based & 2025 & 100\,m \\
\bottomrule
\end{tabular}

\vspace{1mm}
\parbox{\columnwidth}{\footnotesize
\textit{Note:} WSF Tracker used \texttt{as\_of\_code=19} as its common baseline cutoff. The 2025 GHSL layer represents a projected epoch.
}
\end{table}

\subsection{Raster Quantity and Allocation Disagreement}
\label{subsec:supp_quantity_allocation}

Quantity disagreement dominated raster error at 100\,m, accounting for 83--99\% of total disagreement, with median allocation disagreement below 0.016 (Table~\ref{tab:supp_quantity_allocation}). At 10\,m, allocation disagreement contributed 28--31\%, indicating greater sensitivity to positional differences at finer spatial support. Thus, agreement in settlement location did not necessarily imply accurate estimation of built-up quantities.

\begin{table}[!ht]
\centering
\caption{Median per-tile quantity disagreement (QD), allocation disagreement (AD), and quantity share of total disagreement.}
\label{tab:supp_quantity_allocation}
\renewcommand{\arraystretch}{1.15}
\setlength{\tabcolsep}{5pt}
\small

\begin{tabular}{lcccc}
\toprule
\textbf{Product}
& \textbf{Grid}
& \textbf{QD}
& \textbf{AD}
& \textbf{QD share} \\
\midrule
WSF Tracker & 100\,m & 0.066 & 0.014 & 83\% \\
GHSL        & 100\,m & 0.147 & 0.015 & 91\% \\
TEMPO       & 100\,m & 0.173 & 0.003 & 98\% \\
OBT         & 100\,m & 0.158 & 0.002 & 99\% \\
\midrule
WSF Tracker & 10\,m & 0.081 & 0.037 & 69\% \\
OBT         & 10\,m & 0.066 & 0.025 & 72\% \\
\bottomrule
\end{tabular}
\end{table}

\subsection{Building-Size and Count--Area Diagnostics}
\label{subsec:supp_small_buildings}

\begin{figure*}
\centering
\includegraphics[width=\linewidth]{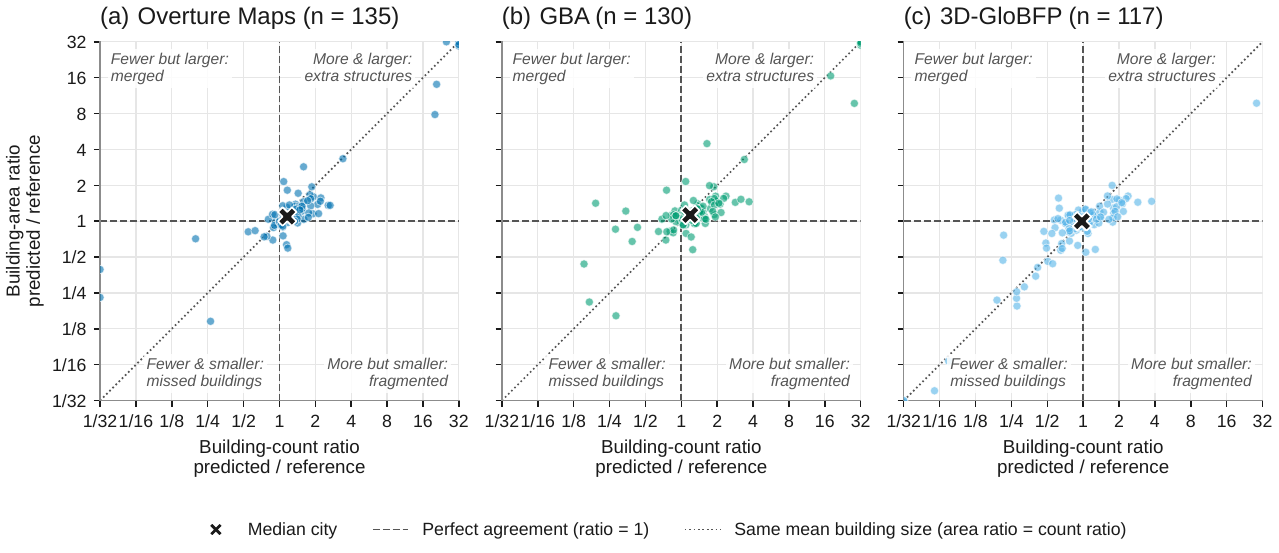}
\caption{City-level building-count versus building-area ratios (predicted/reference; logarithmic axes). Dashed lines indicate perfect agreement and the diagonal equal proportional change. Points below the diagonal represent comparatively more but smaller predicted footprints; points above indicate fewer but larger footprints.}
\label{fig:supp_count_area}
\end{figure*}

City-level comparisons showed greater deviations in building-count ratios than in total-area ratios, particularly for Overture and GBA (Figure~\ref{fig:supp_count_area}). This pattern is consistent with excess small footprints or fragmentation, whereas proportional underestimation of counts and area suggests incomplete detection. Extreme ratios in a few study areas may also reflect incomplete reference coverage.

\begin{figure*}
\centering
\includegraphics[width=\linewidth]{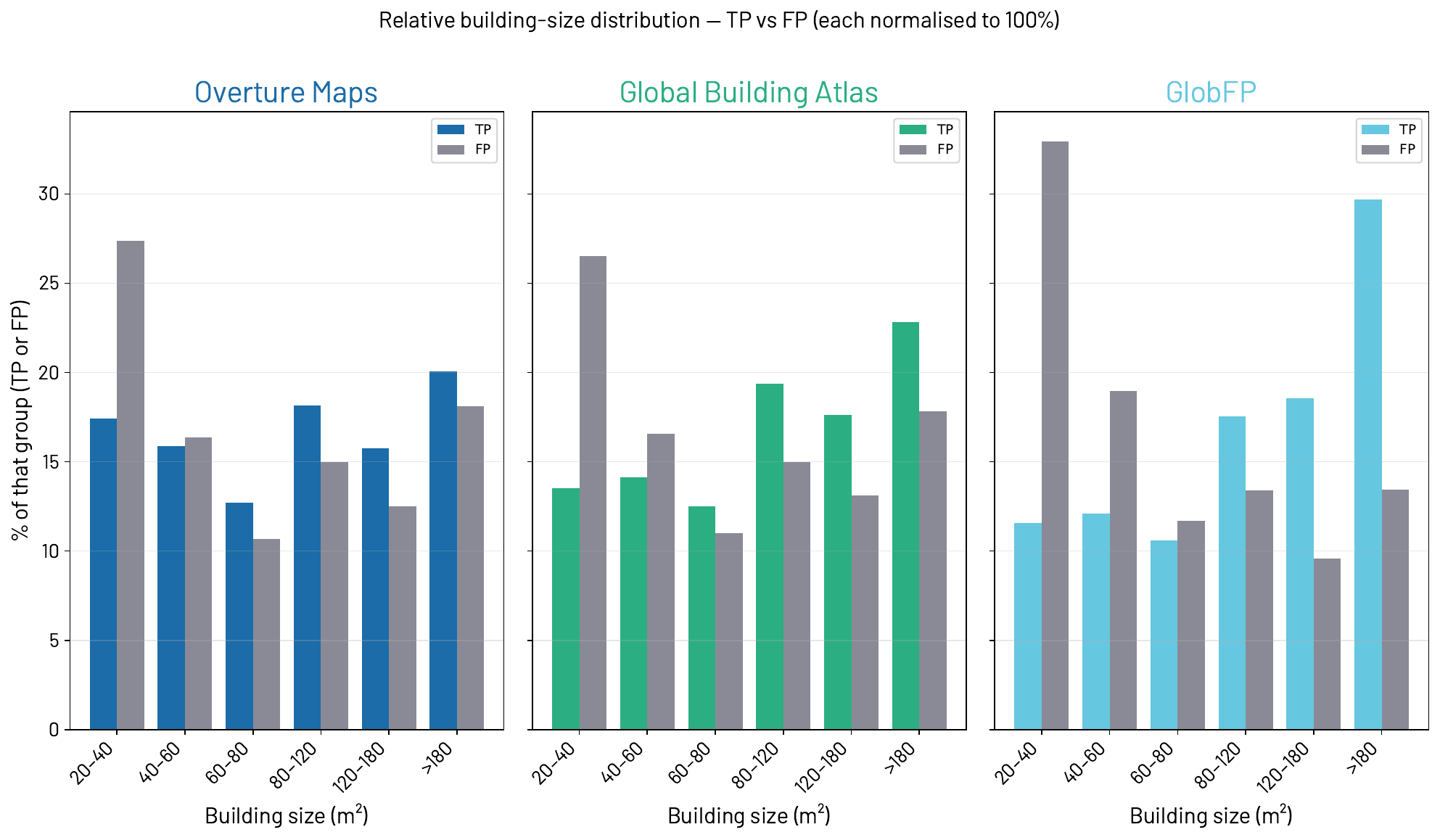}
\caption{Relative size distributions of true-positive (TP) and false-positive (FP) vector footprints, normalized separately within each product. False positives are concentrated among smaller footprints.}
\label{fig:supp_small_buildings}
\end{figure*}

False-positive footprints were consistently smaller than matched true positives (Table~\ref{tab:supp_small_buildings}). In the 20--40\,m$^2$ class, 71--81\% of candidate footprints were false positives, with the strongest concentration of small-footprint errors observed for GBA and 3D-GloBFP (Figure~\ref{fig:supp_small_buildings}).

\begin{table}[!ht]
\centering
\caption{Median footprint areas by detection class and false-positive share in the 20--40\,m$^2$ class.}
\label{tab:supp_small_buildings}
\renewcommand{\arraystretch}{1.15}
\setlength{\tabcolsep}{4pt}
\small

\begin{tabular}{lcccc}
\toprule
\textbf{Dataset}
& \textbf{TP}
& \textbf{FP}
& \textbf{FN}
& \textbf{FP share} \\
\midrule
Overture Maps & 87.4  & 71.2 & 87.3 & 71\% \\
GBA           & 98.4  & 71.6 & 62.5 & 81\% \\
3D-GloBFP     & 115.5 & 57.7 & 75.4 & 78\% \\
\bottomrule
\end{tabular}

\vspace{1mm}
\parbox{\columnwidth}{\footnotesize
\textit{Note:} TP, FP, and FN values are median footprint areas in m$^2$.
}
\end{table}

\subsection{Temporal-Alignment Sensitivity}
\label{subsec:supp_temporal_alignment}

Reevaluating WSF Tracker using the six-month observation closest to each reference-imagery date increased city-level F1 by a mean of 0.060 (median = 0.037; maximum = 0.351). Later observations generally increased recall but reduced precision because newly developed settlements were counted as false positives against older references. Temporal mismatch may therefore contribute to apparent errors in rapidly changing settlements.

\end{document}